\def\arxiv{1}

\ifx\arxiv\undefined
\documentclass[journal,twoside,web]{ieeecolor}
\else
\documentclass[journal,print]{ieeecolor}
\fi
\usepackage{cite}
\usepackage{amsmath,amssymb,amsfonts}
\usepackage{algorithmic}
\usepackage{graphicx}
\usepackage{textcomp}

\usepackage{mathtools}
\usepackage{subcaption}
\usepackage{algorithm}
\usepackage{siunitx}
\usepackage{tikz}

\usetikzlibrary{positioning}
\usetikzlibrary{shapes.geometric}

\ifx\arxiv\undefined
\def\logoname{LOGO-generic-web}
\definecolor{subsectioncolor}{rgb}{0,0.541,0.855}
\def\journalname{Generic Colorized Journal}
\else
\def\logoname{LOGO-generic-web}
\definecolor{subsectioncolor}{rgb}{0,0,0}
\definecolor{nblue}{rgb}{0,0,0}
\def\journalname{Journal}
\fi

\newtheorem{theorem}{Theorem}

\newtheorem{lemma}{Lemma}

\newtheorem{definition}{Definition}
\newtheorem{problem}{Problem}

\newtheorem{example}{Example}

\ifx\arxiv\undefined
\newcommand{\update}[1]{{\color{red}#1}}
\else
\newcommand{\update}[1]{#1}
\fi

\def\BibTeX{{\rm B\kern-.05em{\sc i\kern-.025em b}\kern-.08em
    T\kern-.1667em\lower.7ex\hbox{E}\kern-.125emX}}
\begin{document}
\title{Learning Optimal Strategies for Temporal Tasks\\ in Stochastic Games}
\author{Alper Kamil Bozkurt, Yu Wang, Michael M. Zavlanos, Miroslav Pajic
\thanks{Manuscript received March 24, 2022; revised August 30, 2023. This work is sponsored in part by the ONR under agreements N00014-17-1-2504 and N00014-20-1-2745, AFOSR under award number FA9550-19-1-0169, and the NSF under CNS-1652544 and CNS-1932011 award, and the National AI Institute for Edge Computing Leveraging Next Generation Wireless Networks, Grant CNS-2112562. A very preliminary version of some of the results appeared in~\cite{bozkurt2021a}.}
\thanks{A. K. Bozkurt is with the Department of Computer Science, Duke University, Durham, NC 27708, USA (e-mail: alper.bozkurt@duke.edu).}
\thanks{Y. Wang is with the Department of Mechanical and Aerospace Engineering, University of Florida, Gainesville, FL 32611, USA (e-mail: yuwang1@ufl.edu).}
\thanks{M. M. Zavlanos is with the Department of Mechanical Engineering and Material Science, Duke University, Durham, NC 27708, USA (e-mail: mz61@duke.edu).}
\thanks{M. Pajic is with the Department of Electrical and Computer Engineering, Duke University, Durham, NC 27708, USA (e-mail: miroslav.pajic@duke.edu).}
}

\maketitle

\begin{abstract}
Synthesis from linear temporal logic~(LTL) specifications provides assured controllers for %
systems operating in stochastic and potentially adversarial environments. 
Automatic synthesis tools, however, require a model of the environment to construct controllers.
In this work, we introduce a \textit{model-free} reinforcement learning (RL) approach to derive controllers from given LTL specifications even when the environment is completely unknown.
We model the problem %
as a stochastic game (SG) between the controller and the adversarial environment; we then learn optimal control strategies that maximize the probability of satisfying the LTL specifications against the \textit{worst-case} environment behavior.
We first construct a product game using the deterministic parity automaton (DPA) translated from the given LTL specification. 
By deriving distinct rewards and discount factors from the acceptance condition of the DPA, we reduce the maximization of the worst-case probability of satisfying the LTL specification into the maximization of a discounted reward objective in the product game; this enables the use of model-free RL algorithms to learn an optimal controller strategy.
To deal with the common scalability problems when the number of \update{sets defining the acceptance condition of the DPA (usually referred as colors),} is large, we propose a lazy color generation method where distinct rewards and discount factors are utilized only when needed, and an approximate method where the controller eventually focuses on only one color.
In several case studies, we show that our approach is scalable to a wide range of LTL formulas, significantly outperforming existing methods for learning controllers from LTL specifications in~SGs. %
\end{abstract}


\section{Introduction} \label{sec:intro}

Linear temporal logic (LTL)~\cite{baier2008} provides a formal specification language that can be used to express many control tasks  with temporal objectives, such as safety, sequencing, conditioning, and repetition. 
For instance, majority of robotics tasks can be expressed as an LTL formula (e.g., \cite{kloetzer2007,kress2009,kloetzer2010,chen2012,lahijanian2012,ding2014}) such as %
path planning task 
where the objective includes %
``first go to the entrance and then go to the workspace'', ``if the battery is low, go to the charger'', and ``continuously monitor a particular region while avoiding the danger zones''.~Due~to~this expressiveness, automatic controller synthesis from LTL specifications has been %
widely studied in the control community 
(e.g., \cite{fainekos2009,kress2009,kloetzer2010,chen2012,lahijanian2012,ding2014,jiang2004,tabuada2006,kloetzer2008,wongpiromsarn2012,yordanov2012,liu2013,zamani2014}).
However, such synthesis approaches %
require a model operational environment can be specified a priori, which is mostly impracticable as modern autonomous systems are being deployed in more complex and stochastic environments.

The common situations where a model of the environment is not available necessitate the use of learning-based methods to design controllers for LTL specifications.
A growing body of literature has investigated learning control strategies %
when the interaction with the stochastic environment can be modeled as a Markov decision process~(MDP). 
Several studies (e.g., \cite{fu2014a,brazdil2014,gao2020}) have introduced model-based probably approximately correct (PAC) methods that construct a product MDP using the deterministic Rabin automata (DRAs) derived from the~LTL specifications and detect the product MDP components where the specifications are satisfied.
However, to detect such components, these methods learn and store the transition structure of the MDP, potentially resulting in large memory requirements.

Recently, model-free reinforcement learning (RL) methods have been proposed to mitigate this problem (e.g., \cite{hahn2019,bozkurt2020}). With such methods, the LTL specifications are translated into suitable limit-deterministic B\"uchi automata (LDBAs) \cite{hahn2015} to construct a product MDP. Rewards are then derived from the B\"uchi (i.e., repeated reachability) acceptance condition such that control strategies maximizing sums of discounted rewards maximize the satisfaction probabilities; then, off-the-shelf RL algorithms are used to learn such strategies.
However, all these approaches focus on learning for MDPs and do not consider \emph{nondeterministic}
adversarial environments that can take non-random actions to disrupt performing the given~tasks.

A key challenge in enabling autonomous systems to operate in unpredictable environments is to learn control strategies robust to adversarial inputs. The adversarial inputs usually cannot be restricted to simple perturbations, and thus they need to be considered as separate unpredictable environment actions (e.g., \cite{liu2013,fawzi2014}).  \update{The design of controllers resilient against the worst-case environment behavior is often crucial for autonomy when safety and security are of the utmost importance. Though such controllers might be conservative, they provide assurance against any, potentially adversarial, environment behaviors.}

Formally, the interaction between controllers and potentially adversarial environments can be modeled as zero-sum turn-based stochastic games (SGs)~\cite{chatterjee2012} where the objective is to learn optimal controller strategies to perform the tasks specified as LTL formulas.
SGs are natural extensions of MDPs to systems where some states are under control of an adversary whose objective is to prevent performing the given tasks. \update{Although more restrictive than the general-sum~concurrent games, the zero-sum turn-based games %
are used to model
many} control and sequential decision-making problems where tasks need to be successfully carried out no matter how the adversarial environment acts~\cite{filar1997,neyman2003}; \update{thus, are widely adopted by the control community (e.g., \cite{wen2021,niu2020,niu2021,zhang2021,fu2014b,fu2016,svorevnova2016})}. For example, the problem of security-aware motion planning against stealthy attacks can be expressed as an SG~\cite{elfar19a, elfar19b, bozkurt2021b}.

\update{There have been studies investigating controller synthesis from LTL specifications in SGs \cite{svorevnova2016} including synthesis from Generalized Reactivity(1) fragment of LTL \cite{svorevnova2017} and adaptive approaches \cite{sadraddini2017}; yet, only few have investigated} the use of learning for LTL specifications in SGs.
A model-based PAC method from~\cite{wen2021} pre-computes the winning states with respect to the LTL specification based on the assumption that the given LTL specifications can be translated into deterministic B\"uchi automata (DBAs) and the transition structure of the SG is available \emph{a-priori}. Another model-based PAC method from~\cite{ashok2019} learns the winning states for reachability specifications, a limited fragment of LTL that cannot be used to specify continuous tasks. As model-based methods, these are inefficient in terms of memory requirements when the number of different states that actions can lead to is large.

Recently, two automata-based model-free learning methods have been proposed to learn control strategies for LTL specifications in SGs~\cite{hahn2020,bozkurt2021a}. 
Our preliminary work~\cite{bozkurt2021a} has introduced a method that translates the LTL specifications into DRAs, and provides reward and discount factors for each Rabin pair in the acceptance conditions. This method, however, does not guarantee convergence to the optimal strategies when there is more than one Rabin pair.
In~\cite{hahn2020}, the LTL specifications are translated to deterministic parity automata (DPAs), whose acceptance conditions are more intricate than the B\"uchi condition.
The problem of satisfying the parity condition is then reduced to a reachability problem by extending the product game constructed using DPAs with terminal states such that the probabilities of transitions to these states are determined by the sets in the parity condition.~Yet,~the~length of the episodes required to learn optimal strategies in this method \emph{grows exponentially with the number of sets};~thus,~substantially limiting its scalability as we show in our case studies.

Consequently, in this work, we introduce an approach to learn optimal strategies for \emph{any} desired LTL specifications in SGs while significantly improving scalability compared to the existing methods. 
Our approach is \emph{model-free}; i.e., it does not use or construct transition models of SGs.
We start by composing an SG where the transition structure is completely unknown with a DPA that is automatically obtained from the given LTL specification. 
We then derive novel reward and discount factors from the parity acceptance condition such that a strategy maximizing the sum of the discounted rewards also \emph{maximizes the probability of satisfying the LTL specification in the worst case} -- i.e., for the worst (most-damaging) adversary actions from the controller's perspective. \update{To overcome the scalability issues when the number of sets defining the acceptance condition of the DPA, referred to as \emph{colors}, is large}, we provide a model-free method that uses these distinct rewards and discount factors \emph{only when needed} via lazy color generation and thus facilitates efficient learning of such optimal strategies in the derived product SGs. 

We also provide a scalable approximate method for the scenarios where the controller can perform the task by eventually focusing on only one color.
Finally, we compare our methods with each other, and with the existing methods in several robot navigation case studies with the LTL specifications, and we show that our lazy color generation and approximate methods outperform the others. Furthermore, we demonstrate the applicability of our methods in several robotic arm tasks.

The rest of the paper is organized as follows. Section~\ref{sec:prelim} reviews necessary preliminaries, before providing a reduction from LTL specifications to reward returns in Section~\ref{sec:reduction}. To improve learning scalability, we introduce the lazy color generation framework in Section~\ref{sec:lazy}, and evaluate our approach in Section~\ref{sec:case}. Finally, Section~\ref{sec:conc} provides concluding~remarks.

\section{Preliminaries and Problem Formulation} \label{sec:prelim}

\subsection{Stochastic Games}
We use SGs to model the problem of 
performing a given task by the controller (Player~1) against an adversary (Player~2) in a stochastic environment.

\begin{definition}
A (labeled \update{fully-observable} turn-based two-player) \textbf{stochastic game} is a tuple $\mathcal{G}=(S, (S_\mu, S_\nu), s_0,\allowbreak A, P, \textnormal{AP}, L)$ where 
$S=S_\mu \cup S_\nu$ is a finite set of states, $S_\mu$ is the set of states in which the controller takes action, $S_\nu$ is the set of states under the control of the adversary, and $s_0$ is the initial state;
$A$ is a finite set of actions and $A(s)$ denotes the set of actions that can be taken in a state $s$; $P:S\times A\times S\mapsto [0,1]$ is a probabilistic transition function such that $\sum_{s'\in S} P(s,a,s')=1$ if $a \in A(s)$ and $0$ otherwise; finally, 
\textnormal{AP} is a finite set of atomic propositions and $L:S\mapsto 2^\textnormal{AP}$ is a labeling~function.
\end{definition}

SGs can be considered as games played by the controller and the adversary for infinitely many time steps on finite directed graphs consisting of state and state-action nodes. \update{The state nodes are divided into two distinct subsets reflecting the turns of the controller and the adversary. The game starts in the initial state and moves between the state nodes as the controller and the adversary take actions.} In each state node, only the owner of the state \update{observes the state information and} chooses one of the state-action nodes of the state and the game then probabilistically transitions to one of the successors of the node according to the given transition function.

\begin{example} \label{ex:sg}
An example SG of a robotics environment is shown in Fig.~\ref{fig:sg}. A robotic agent (i.e., the controller) starts in ``\textnormal{Entrance}'', where an empty label $\{\}$ is received. 
In ``\textnormal{Entrance}'', the agent can choose either ``\textnormal{go up}'' or ``\textnormal{go down}''. 
With ``\textnormal{go down}'', the agent starts working and moves to the charging station with probability (w.p.) $0.9$ (i.e., the game makes a transition to ``\textnormal{Charger(Agent-On)}'' and a label of $\{charging,working\}$ is received). If the agent keeps taking ``\textnormal{go down}'', it eventually moves to ``\textnormal{Charger(Agent-On)}'' w.p.~$1$. With ``\textnormal{go up}'', the agent moves to the charging station but could get stuck w.p.~$0.1$. In ``\textnormal{Charger(Agent-On)}'', the adversary can prevent the agent from working by turning it off (``\textnormal{Charger(Agent-Off)}''); however, the agent can turn itself on again. In addition, the adversary can move the agent from the charging station to ``\textnormal{Workspace}'', where the agent cannot get charged but can work or go back to the charging~station. %
\end{example}

We call $\pi\coloneqq s_0s_1\dots$ an infinite \emph{path} (i.e., execution) of the SG $\mathcal{G}$ if for all $t\geq0$ there exists an action $a\in A(s_t)$ such that $P(s_t,a,s_{t+1})>0$. We denote the state $s_t$ and the suffix $s_t s_{t+1}...$ by $\pi[t]$ and $\pi[t{:}]$, respectively. The behaviors of the players can be specified by strategy functions mapping the history of the visited states to an action. We focus on finite-memory strategies since they suffice for the LTL tasks (see \cite{chatterjee2012} and references therein).

\begin{figure*}
\centering
\begin{subfigure}[b]{0.59\linewidth} 
    \centering
    \resizebox{\textwidth}{!}{\begin{tikzpicture}[
    controller/.style={circle, draw=black, very thick, minimum size=30pt, inner sep=2pt},
    adversary/.style={rectangle, draw=black, very thick, minimum size=30pt, inner sep=2pt},
    action/.style={diamond, draw=black, very thick, minimum size=30pt, inner sep=1pt},
    probability/.style={rectangle, inner sep=1pt}]
    
    \node[controller]  (Entrance) {\shortstack{Entrance\\\textit{\small \{\}}}};
    \node[action] (EntranceGoUp) at (2,1) {\shortstack{\small go\\\small up}};
    \node[action] (EntranceGoDown) at (2,-1) {\shortstack{\small go\\\small down}};
    \node[probability] (EntranceGoUp01) at (3,1.5) {$0.1$};
    \node[probability] (EntranceGoUp09) at (3,0.5) {$0.9$};
    \node[probability] (EntranceGoDown09) at (3,-0.5) {$0.9$};
    \node[probability] (EntranceGoDown01) at (3,-1.5) {$0.1$};
    
    \node[adversary]  (ChargerStuck) at (4.5,3) {\shortstack{Charger\\(Agent-Stuck)\\\textit{\small \{charging\}}}};
    \node[action] (ChargerStuckStay) at (6.5,3) {\shortstack{\small stay}};
    \node[probability] (ChargerStuckStay10) at (7.5,3) {$1.0$};
    
    \node[adversary]  (ChargerOn) at (4.5,0) {\shortstack{Charger\\(Agent-On)\\\textit{\small \{charging,}\\\textit{\small working\}}}};
    \node[action] (ChargerOnTurnOff) at (6.5,1) {\shortstack{\small turn \\\small off}};
    \node[action] (ChargerOnMove) at (6.5,-1.5) {\shortstack{\small move}};
    \node[probability] (ChargerOnTurnOff10) at (7.5,1) {$1.0$};
    \node[probability] (ChargerOnMove10) at (7.5,-1.5) {$1.0$};
    
    \node[controller]  (ChargerOff) at (9.5,1) {\shortstack{Charger\\(Agent-Off)\\\textit{\small \{charging\}}}};
    \node[action] (ChargerOffTurnOn) at (11.5,1) {\shortstack{\small turn \\\small on}};
    \node[probability] (ChargerOffTurnOn10) at (12.5,1) {$1.0$};
    
    \node[controller]  (Workspace) at (9.5,-1.5) {\shortstack{Workspace\\\textit{\small \{working\}}}};
    \node[action] (WorkspaceGoBack) at (11.5,-1.5) {\shortstack{\small go \\\small back}};
    \node[probability] (WorkspaceGoBack10) at (12.5,-1.5) {$1.0$};
    
    \draw[->, very thick] (-1.5,0) -- (Entrance.west);
    \draw[->, very thick] (Entrance.east) -- (EntranceGoUp.west);
    \draw[->, very thick] (Entrance.east) -- (EntranceGoDown.west);
    
    \draw[very thick] (EntranceGoUp.east)  .. controls +(right:5pt) and +(down:5pt).. (EntranceGoUp01.south);
    \draw[->, very thick] (EntranceGoUp01.north)  .. controls +(up:5pt) and +(left:20pt).. (ChargerStuck.west);
    \draw[very thick] (EntranceGoUp.east)  .. controls +(right:5pt) and +(up:5pt).. (EntranceGoUp09.north);
    \draw[->, very thick] (EntranceGoUp09.south)  .. controls +(down:5pt) and +(left:5pt).. (ChargerOn.west);
    
    \draw[very thick] (EntranceGoDown.east)  .. controls +(right:5pt) and +(down:5pt).. (EntranceGoDown09.south);
    \draw[->, very thick] (EntranceGoDown09.north)  .. controls +(up:5pt) and +(left:5pt).. (ChargerOn.west);
    \draw[very thick] (EntranceGoDown.east)  .. controls +(right:5pt) and +(up:5pt).. (EntranceGoDown01.north);
    \draw[->, very thick] (EntranceGoDown01.south)  .. controls +(down:10pt) and +(down:40pt).. (Entrance.south);
    
    \draw[->, very thick] (ChargerStuck.east) -- (ChargerStuckStay.west);
    \draw[very thick] (ChargerStuckStay.east) -- (ChargerStuckStay10.west);
    \draw[->, very thick] (ChargerStuckStay10.north)  .. controls +(up:40pt) and +(up:20pt).. (ChargerStuck.north);
    
    \draw[->, very thick] (ChargerOn.east) -- (ChargerOnTurnOff.west);
    \draw[very thick] (ChargerOnTurnOff.east) -- (ChargerOnTurnOff10.west);
    \draw[->, very thick] (ChargerOnTurnOff10.east)  -- (ChargerOff.west);
    \draw[->, very thick] (ChargerOn.east) -- (ChargerOnMove.west);
    \draw[very thick] (ChargerOnMove.east) -- (ChargerOnMove10.west);
    \draw[->, very thick] (ChargerOnMove10.east)  -- (Workspace.west);
    
    \draw[->, very thick] (ChargerOff.east) -- (ChargerOffTurnOn.west);
    \draw[very thick] (ChargerOffTurnOn.east) -- (ChargerOffTurnOn10.west);
    \draw[->, very thick] (ChargerOffTurnOn10.north)  .. controls +(up:60pt) and +(up:30pt).. (ChargerOn.north);
    
    \draw[->, very thick] (Workspace.east) -- (WorkspaceGoBack.west);
    \draw[very thick] (WorkspaceGoBack.east) -- (WorkspaceGoBack10.west);
    \draw[->, very thick] (WorkspaceGoBack10.south)  .. controls +(down:60pt) and +(down:40pt).. (ChargerOn.south);
    
    \node[draw,circle,draw=black, very thick, inner sep=5pt] (controller_shape) at (-1,4) {};
    \node[draw,rectangle,draw=black, very thick, inner sep=6pt] (adversary_shape) at (-1,3.3) {};
    \node[draw,diamond,draw=black, very thick, inner sep=4pt] (action_shape) at (-1,2.6) {};
    \node[draw=white,fill=white,inner sep=1pt] (prob) at (-1,1.9) {p};
    \node[] (controller_text) at (0.8,4.03) {: Controller States};
    \node[] (adversary_text) at (0.8,3.3) {: Adversary States};
    \node[] (action_text) at (0.1,2.63) {: Actions};
    \node[] (prob_text) at (0.6,1.9) { \shortstack{Probabilistic\\Transitions}};
    \node[] (prob_text_) at (-0.5,1.9) {:};
    \node[] (prob_in) at (-1.5,1.9) {};
    \node[] (prob_out) at (-0.7,1.9) {};
    \draw[very thick] (prob_in.east) -- (prob.west);
    \draw[->, very thick] (prob.east) -- (prob_out.east);
    \end{tikzpicture}}
    \vspace{-32pt}
    \caption{An example SG. The sets of words under the state names are the labels.}
    \label{fig:sg}
\end{subfigure}
\hfill
\begin{subfigure}[b]{0.38\linewidth} 
    \centering
    \resizebox{\textwidth}{!}{
    \begin{tikzpicture}[
    q/.style={circle, draw=black, very thick, minimum size=30pt, inner sep=2pt},
    label/.style={rectangle, inner sep=1pt}]
    
    \node[q]  (q0) {\shortstack{{\Large $q_0$}}};
    \node[label] (q0cw) at (0.5,3.5) {\shortstack{$(1)$\small \textit{\{charging,working\}}}};
    \node[label] (q0w) at (0.5,2.5) {\shortstack{$(0)$\small \textit{\{working\}}}};
    \node[label] (q0e) at (0.5,1.5) {\shortstack{$(2)$\small \textit{\{\}}}};
    \node[label] (q0c) at (3,0) {\shortstack{$(2)$\small \textit{\{charging\}}}};
    
    \node[q]  (q1) at (5.5,0) {\shortstack{{\Large $q_1$}}};
    \node[label] (q1ew) at (3,-1) {\shortstack{$(2)$\small\textit{\{\}}$\mid$\\\small\textit{\{working\}}}};
    \node[label] (q1ccw) at (5.5,2) {\shortstack{$(1)$\small\textit{\{charging\}}$\mid$\\\small \textit{\{charging,working\}}}};
    
    \draw[->, very thick] (0,1) -- (q0.north);
    \draw[very thick] (q0.east)..controls+(right:50pt)and+(right:20pt).. (q0cw.east);
    \draw[->, very thick] (q0cw.west)..controls+(left:20pt)and+(left:30pt).. (q0.west);
    \draw[very thick] (q0.east)..controls+(right:45pt)and+(right:15pt).. (q0w.east);
    \draw[->, very thick] (q0w.west)..controls+(left:30pt)and+(left:20pt).. (q0.west);
    \draw[very thick] (q0.east)..controls+(right:25pt)and+(right:15pt).. (q0e.east);
    \draw[->, very thick] (q0e.west)..controls+(left:30pt)and+(left:10pt).. (q0.west);
    \draw[very thick] (q0.east) -- (q0c.west);
    \draw[->, very thick] (q0c.east) -- (q1.west);
    
    \draw[very thick] (q1.east)..controls+(right:30pt)and+(right:60pt).. (q1ew.east);
    \draw[->, very thick] (q1ew.west)..controls+(left:80pt)and+(left:30pt).. (q0.west);
    \draw[very thick] (q1.east)..controls+(right:40pt)and+(right:10pt).. (q1ccw.east);
    \draw[->, very thick] (q1ccw.west)..controls+(left:10pt)and+(left:30pt).. (q1.west);
    \end{tikzpicture}}
    \caption{A DPA derived from LTL formula $\varphi=(\lozenge \square \textit{working} \wedge \square \lozenge \textit{charging}) \vee \lozenge \square \textit{charging}$. %
    The numbers within parentheses and the set of words on the transitions are the colors and the labels of the transitions, respectively.}
    \label{fig:dpa}
\end{subfigure}
\begin{subfigure}[b]{\linewidth} 
    \centering
    \resizebox{\textwidth}{!}{\begin{tikzpicture}[
       controller/.style={circle, draw=black, very thick, minimum size=30pt, inner sep=2pt},
       adversary/.style={rectangle, draw=black, very thick, minimum size=30pt, inner sep=2pt},
       action/.style={diamond, draw=black, very thick, minimum size=30pt, inner sep=1pt},
       probability/.style={rectangle, inner sep=1pt}]
       
       \node[controller]  (Entrance0q) {\shortstack{Entrance\\\large $q_0$,\normalsize $(2)$}};
       \node[action] (Entrance0qGoUp) at (2,1) {\shortstack{\small go\\\small up}};
       \node[action] (Entrance0qGoDown) at (2,-1) {\shortstack{\small go\\\small down}};
       \node[probability] (Entrance0qGoUp01) at (3,1.5) {$0.1$};
       \node[probability] (Entrance0qGoUp09) at (3,0.5) {$0.9$};
       \node[probability] (Entrance0qGoDown09) at (3,-0.5) {$0.9$};
       \node[probability] (Entrance0qGoDown01) at (3,-1.5) {$0.1$};
       
       \node[adversary]  (ChargerStuck0q) at (4.5,3.5) {\shortstack{Charger\\(Agent-Stuck)\\\large $q_0$,\normalsize $(2)$}};
       \node[action] (ChargerStuck0qStay) at (6.5,3.5) {\shortstack{\small stay}};
       \node[probability] (ChargerStuck0qStay10) at (7.5,3.5) {$1.0$};
       
       \node[adversary]  (ChargerStuck1q) at (9.5,3.5) {\shortstack{Charger\\(Agent-Stuck)\\\large $q_1$,\normalsize $(1)$}};
       \node[action] (ChargerStuck1qStay) at (11.5,3.5) {\shortstack{\small stay}};
       \node[probability] (ChargerStuck1qStay10) at (12.5,3.5) {$1.0$};
       
       \node[adversary]  (ChargerOn0q) at (4.5,0) {\shortstack{Charger\\(Agent-On)\\\large $q_0$,\normalsize $(1)$}};
       \node[action] (ChargerOn0qTurnOff) at (6.5,1.5) {\shortstack{\small turn \\\small off}};
       \node[action] (ChargerOn0qMove) at (6.5,-1) {\shortstack{\small move}};
       \node[probability] (ChargerOn0qTurnOff10) at (7.5,1.5) {$1.0$};
       \node[probability] (ChargerOn0qMove10) at (7.5,-1) {$1.0$};
       
       \node[adversary]  (ChargerOn1q) at (14,1.5) {\shortstack{Charger\\(Agent-On)\\\large $q_1$,\normalsize $(1)$}};
       \node[action] (ChargerOn1qTurnOff) at (16,2.5) {\shortstack{\small turn \\\small off}};
       \node[action] (ChargerOn1qMove) at (16,0) {\shortstack{\small move}};
       \node[probability] (ChargerOn1qTurnOff10) at (17,2.5) {$1.0$};
       \node[probability] (ChargerOn1qMove10) at (17,0) {$1.0$};
       
       \node[controller]  (ChargerOff0q) at (9.5,1.5) {\shortstack{Charger\\(Agent-Off)\\\large $q_0$,\normalsize $(2)$}};
       \node[action] (ChargerOff0qTurnOn) at (11.5,1.5) {\shortstack{\small turn \\\small on}};
       \node[probability] (ChargerOff0qTurnOn10) at (12.5,1.5) {$1.0$};
       
       \node[controller]  (ChargerOff1q) at (19,2.5) {\shortstack{Charger\\(Agent-Off)\\\large $q_1$,\normalsize $(1)$}};
       \node[action] (ChargerOff1qTurnOn) at (21,2.5) {\shortstack{\small turn \\\small on}};
       \node[probability] (ChargerOff1qTurnOn10) at (22,2.5) {$1.0$};
       
       \node[controller]  (Workspace0q) at (9.5,-1) {\shortstack{Workspace\\\large $q_0$,\normalsize $(0)$}};
       \node[action] (Workspace0qGoBack) at (11.5,-1) {\shortstack{\small go \\\small back}};
       \node[probability] (Workspace0qGoBack10) at (12.5,-1) {$1.0$};
       
       \node[controller]  (Workspace1q) at (19,0) {\shortstack{Workspace\\\large $q_1$,\normalsize $(2)$}};
       \node[action] (Workspace1qGoBack) at (21,0) {\shortstack{\small go \\\small back}};
       \node[probability] (Workspace1qGoBack10) at (22,0) {$1.0$};

       \draw[->, very thick] (-1.5,0) -- (Entrance0q.west);
       \draw[->, very thick] (Entrance0q.east) -- (Entrance0qGoUp.west);
       \draw[->, very thick] (Entrance0q.east) -- (Entrance0qGoDown.west);
       
       \draw[very thick] (Entrance0qGoUp.east)  .. controls +(right:5pt) and +(down:5pt).. (Entrance0qGoUp01.south);
       \draw[->, very thick] (Entrance0qGoUp01.north)  .. controls +(up:5pt) and +(left:20pt).. (ChargerStuck0q.west);
       \draw[very thick] (Entrance0qGoUp.east)  .. controls +(right:5pt) and +(up:5pt).. (Entrance0qGoUp09.north);
       \draw[->, very thick] (Entrance0qGoUp09.south)  .. controls +(down:5pt) and +(left:5pt).. (ChargerOn0q.west);
       
       \draw[very thick] (Entrance0qGoDown.east)  .. controls +(right:5pt) and +(down:5pt).. (Entrance0qGoDown09.south);
       \draw[->, very thick] (Entrance0qGoDown09.north)  .. controls +(up:5pt) and +(left:5pt).. (ChargerOn0q.west);
       \draw[very thick] (Entrance0qGoDown.east)  .. controls +(right:5pt) and +(up:5pt).. (Entrance0qGoDown01.north);
       \draw[->, very thick] (Entrance0qGoDown01.south)  .. controls +(down:10pt) and +(down:40pt).. (Entrance0q.south);
       
       \draw[->, very thick] (ChargerStuck0q.east) -- (ChargerStuck0qStay.west);
       \draw[very thick] (ChargerStuck0qStay.east) -- (ChargerStuck0qStay10.west);
       \draw[->, very thick] (ChargerStuck0qStay10.east)  -- (ChargerStuck1q.west);
       
       \draw[->, very thick] (ChargerStuck1q.east) -- (ChargerStuck1qStay.west);
       \draw[very thick] (ChargerStuck1qStay.east) -- (ChargerStuck1qStay10.west);
       \draw[->, very thick] (ChargerStuck1qStay10.north)  .. controls +(up:40pt) and +(up:20pt).. (ChargerStuck1q.north);
       
       \draw[->, very thick] (ChargerOn0q.east) -- (ChargerOn0qTurnOff.west);
       \draw[very thick] (ChargerOn0qTurnOff.east) -- (ChargerOn0qTurnOff10.west);
       \draw[->, very thick] (ChargerOn0qTurnOff10.east)  -- (ChargerOff0q.west);
       \draw[->, very thick] (ChargerOn0q.east) -- (ChargerOn0qMove.west);
       \draw[very thick] (ChargerOn0qMove.east) -- (ChargerOn0qMove10.west);
       \draw[->, very thick] (ChargerOn0qMove10.east)  -- (Workspace0q.west);
       
       \draw[->, very thick] (ChargerOn1q.east) -- (ChargerOn1qTurnOff.west);
       \draw[very thick] (ChargerOn1qTurnOff.east) -- (ChargerOn1qTurnOff10.west);
       \draw[->, very thick] (ChargerOn1qTurnOff10.east)  -- (ChargerOff1q.west);
       \draw[->, very thick] (ChargerOn1q.east) -- (ChargerOn1qMove.west);
       \draw[very thick] (ChargerOn1qMove.east) -- (ChargerOn1qMove10.west);
       \draw[->, very thick] (ChargerOn1qMove10.east)  -- (Workspace1q.west);
       
       \draw[->, very thick] (ChargerOff0q.east) -- (ChargerOff0qTurnOn.west);
       \draw[very thick] (ChargerOff0qTurnOn.east) -- (ChargerOff0qTurnOn10.west);
       \draw[->, very thick] (ChargerOff0qTurnOn10.east) -- (ChargerOn1q.west);
       
       \draw[->, very thick] (ChargerOff1q.east) -- (ChargerOff1qTurnOn.west);
       \draw[very thick] (ChargerOff1qTurnOn.east) -- (ChargerOff1qTurnOn10.west);
       \draw[->, very thick] (ChargerOff1qTurnOn10.north)  .. controls +(up:40pt) and +(up:60pt).. (ChargerOn1q.north);
       
       \draw[->, very thick] (Workspace0q.east) -- (Workspace0qGoBack.west);
       \draw[very thick] (Workspace0qGoBack.east) -- (Workspace0qGoBack10.west);
       \draw[->, very thick] (Workspace0qGoBack10.south)  .. controls +(down:50pt) and +(down:40pt).. (ChargerOn0q.south);
       
       \draw[->, very thick] (Workspace1q.east) -- (Workspace1qGoBack.west);
       \draw[very thick] (Workspace1qGoBack.east) -- (Workspace1qGoBack10.west);
       \draw[->, very thick] (Workspace1qGoBack10.south)  .. controls +(down:80pt) and +(down:80pt).. (ChargerOn0q.south);
       
       \node[draw,circle,draw=black, very thick, inner sep=5pt] (controller_shape) at (-1,4.4) {};
       \node[draw,rectangle,draw=black, very thick, inner sep=6pt] (adversary_shape) at (-1,3.7) {};
       \node[draw,diamond,draw=black, very thick, inner sep=4pt] (action_shape) at (-1,3) {};
       \node[draw=white,fill=white,inner sep=1pt] (prob) at (-1,2.3) {p};
       \node[] (controller_text) at (0.8,4.43) {: Controller States};
       \node[] (adversary_text) at (0.8,3.7) {: Adversary States};
       \node[] (action_text) at (0.1,3.03) {: Actions};
       \node[] (prob_text) at (0.6,2.3) { \shortstack{Probabilistic\\Transitions}};
       \node[] (prob_text_) at (-0.5,2.3) {:};
       \node[] (prob_in) at (-1.5,2.3) {};
       \node[] (prob_out) at (-0.7,2.3) {};
       \draw[very thick] (prob_in.east) -- (prob.west);
       \draw[->, very thick] (prob.east) -- (prob_out.east);
    \end{tikzpicture}}
    \vspace{-32pt}
    \caption{A product game of the SG in \protect\subref{fig:sg} and the DPA in \protect\subref{fig:dpa}. The numbers in the parentheses under the state names are the corresponding colors.}
    \label{fig:prod}
\end{subfigure}
\caption{A product game construction for the system from Example~\ref{ex:sg}.}
\label{fig:graphs}
\end{figure*}

\begin{definition}%
\label{def:strategies}
A \textbf{finite-memory strategy} for an SG $\mathcal{G}$ is a tuple $\sigma=(M,m_0,T,\alpha)$ where $M$ is a finite set of modes; 
$m_0$ is the initial mode; 
$T:M \times S \mapsto \mathcal{D}(M)$ is the transition function that maps the current mode and state to a distribution over the next modes; 
$\alpha:M\times S \mapsto \mathcal{D}(A)$ is a function that maps a given mode $m\in M$ and a state $s\in S$ to a discrete distribution over $A(s)$. 
A \textbf{controller strategy} $\mu$ is a finite-memory strategy that maps only the controller states to distributions over the actions. 
Similarly, an \textbf{adversary strategy} $\nu$ is a finite-memory strategy mapping the adversary states to distributions over the actions. 
A finite-memory strategy is called \textbf{pure memoryless} if there is only one mode ($|M|=1$) and $\alpha(m_0,s)$ is a point distribution assigning a probability of 1 to a single action for all $s\in S$.
\end{definition}

Intuitively, a finite-memory strategy is a finite state machine moving from one mode (memory state) to another as the SG states are visited, outputting a distribution over the actions in each state. 
Unlike the standard definition of finite-memory strategies (e.g.,~\cite{chatterjee2012,baier2008}) where transitions among the modes are all deterministic, Def.~\ref{def:strategies} allows probabilistic transitions; this will later enable modeling of probabilistic transitions between different levels of the derived product~games \update{where pure memoryless strategies induce finite-memory strategies with probabilistic mode transitions in the original SGs}.

For a given pair of a finite-memory controller strategy $\mu$ and an adversary strategy $\nu$ in an SG $\mathcal{G}$, we denote the resulting \emph{induced Markov chain} (MC) as $\mathcal{G}_{\mu,\nu}$. 
We use $\pi \sim \mathcal{G}_{\mu,\nu}$ to denote a path drawn from $\mathcal{G}_{\mu,\nu}$, and we write $\pi_s \sim \mathcal{G}^s_{\mu,\nu}$ to denote a path drawn from the MC $\mathcal{G}^s_{\mu,\nu}$, where $\mathcal{G}^s_{\mu,\nu}$ is same as $\mathcal{G}_{\mu,\nu}$ except that the state $s$ is designated as the initial state instead of $s_0$.
Finally, a \emph{bottom strongly connected component} (BSCC) of an MC is a set of states such that there is a path from each state to any other state in the set without any outgoing transitions. We denote the set of all BSCCs of the MC $\mathcal{G}_{\mu,\nu}$ by $\mathcal{B}(\mathcal{G}_{\mu,\nu})$.

\subsection{Linear Temporal Logic}
LTL provides a high-level formalism to specify tasks with temporal properties by placing requirements for infinite paths \cite{baier2008}. LTL specifications consist of nested combinations of Boolean and temporal operators according to the grammar:
$    \varphi \coloneqq  \mathrm{true} \mid a \mid \varphi_1 \wedge \varphi_2 \mid \neg \varphi \mid \bigcirc \varphi \mid \varphi_1 \textsf{U} \varphi_2, ~ {a\in\textnormal{AP}}.
$ 
The other Boolean operators are defined via the standard equivalences (e.g., $\varphi_1 \vee \varphi_2 \coloneqq \neg(\neg\varphi_1 \wedge \neg\varphi_2)$, $\varphi\to \varphi' \coloneqq \neg \varphi \vee \varphi'$). 
A path $\pi$ of an SG $\mathcal{G}$ satisfies an LTL specification $\varphi$, denoted by $\pi \models \varphi$, if one of the following holds:
\begin{itemize}
\vspace{-2pt}
    \item if $\varphi=a$ and $L(\pi[0])=a$ (i.e., $a$ immediately holds); 
    \item if $\varphi=\varphi_1 \wedge \varphi_2$, $\pi \models \varphi_1$, and $\pi \models \varphi_2$;
    \item if $\varphi=\neg \varphi'$ and $\pi \not\models \varphi'$;
    \item if $\varphi=\bigcirc\varphi'$ (called \emph{next} $\varphi'$) and $\pi[1{:}]\models\varphi'$;
    \item if $\varphi=\varphi_1\textsf{U}\varphi_2$ (called $\varphi_1$ \emph{until} $\varphi_2$) and there exists $t\geq0$ such that $\pi[t{:}]\models \varphi_2$ and for all $0\leq i<t$, $\pi[i{:}]\models \varphi_1$.
\end{itemize}

Intuitively, the temporal operator $\bigcirc \varphi$ expresses that $\varphi$ needs to hold in the next time step, whereas $\varphi_1\textsf{U}\varphi_2$ specifies that $\varphi_1$ needs to hold until $\varphi_2$ holds. 
Other temporal operators such as \emph{eventually} ($\lozenge$) and \emph{always} ($\square$) are also commonly used: 
$\lozenge \varphi \coloneqq \mathrm{true} \ \textsf{U}\ \varphi$ (i.e., $\varphi$ eventually holds) and 
$\square \varphi \coloneqq \neg (\lozenge \neg \varphi)$ (i.e., $\varphi$ always holds).

LTL can be used to specify a wide range of robotics tasks such as sequencing, surveillance, persistence, and avoidance. 
In the following example, we illustrate how a persistence task can be specified as an LTL formula, and we discuss the optimal controller and adversary strategies.

\begin{example} \label{ex:ltl_in_sg}
Consider a simple persistence task ``$\varphi=\lozenge \square \textit{charging}$'' in the SG presented in Fig.~\ref{fig:sg}. 
The robot controller can ensure that ``\textnormal{Charger(Agent-On)}'' is eventually reached by taking ``\textnormal{go down}'' in ``\textnormal{Entrance}'' persistently. However, the adversary can move the robot to ``\textnormal{Workspace}'', a state that is not labeled with ``charging'', thereby making the probability of satisfying the specification $0$. To avoid this, the controller can instead take ``go up'', which w.p. $0.1$ leads to ``\textnormal{Charger(Stuck)}'', a trap cell labeled with ``charging'', where the specification is guaranteed to be satisfied.
\end{example}

\subsection{Deterministic Parity Automata}

Any LTL task can be translated to a DPA that accepts an infinite path satisfying the LTL task \cite{esparza2017}.

\begin{definition}
A \textbf{deterministic parity automaton} is a tuple $\mathcal{A}=(Q, q_0, \Sigma, \delta, \kappa, C)$ such that
$Q$ is the finite set of automaton states;
$q_0 \in Q$ is the initial state;
$\delta: Q \times \Sigma \mapsto Q$ is the transition function;
$\kappa$ is the number of colors;
$C:Q \times \Sigma \mapsto \{0,\dots,\kappa{-}1\}$ is the coloring function.
A \textbf{path} $\pi$ of $\mathcal{G}$ induces an execution $\rho_\pi=\langle q_0, L(\pi[0])\rangle \allowbreak \langle q_1, L(\pi[1])\rangle \dots$ such that for all $t\geq0$, $\delta(q_t,L(\pi[t]))=q_{t+1}$. Let $\textnormal{Inf}(\rho_\pi)$ denote the set of the transitions $\langle q,l\rangle \in Q \times \Sigma$ made infinitely often by $\rho_\pi$; then, a path $\pi$ is accepted by a DPA if $\max\{C(\langle q, l \rangle) \mid \langle q, l \rangle \in \textnormal{Inf}(\rho_\pi)\}$ is an odd number.
\end{definition}

For a given LTL task $\varphi$, we use $\mathcal{A}_\varphi$ to denote a DPA derived from $\varphi$. DPAs provide a systematic way to evaluate the satisfaction of any LTL specification, which can be expressed by the satisfaction of the parity condition of a constructed DPA.
The parity condition is satisfied simply when the largest color among the colors repeatedly visited is an odd number. 
This provides a natural framework to reason about LTL tasks in SGs; the controller tries to visit the states triggering the odd-colored transitions as often as possible, while the adversary tries to do the opposite (i.e., even-colored transitions).

\begin{example}
Fig.~\ref{fig:dpa} shows a DPA derived from the LTL formula ``$\varphi=(\lozenge \square \textit{working} \wedge \square \lozenge \textit{charging}) \vee \lozenge \square \textit{charging}$''.
\update{The executions visiting $q_0$ and $q_1$ infinitely many times make the inter-state transitions colored with 2 infinitely many times. Since 2 is an even number and the largest color in this DPA, regardless of how many times the transitions colored with 1 or 0 are made, these executions would not be accepted. }
Therefore, any accepting execution, after some finite time steps, must stay forever in either $q_0$ or $q_1$.
The executions staying in $q_0$ are accepting only if they receive the label \{charging,working\} infinitely many times and the label \{\} finitely many times, hence satisfying $\lozenge \square \textit{working} \wedge \square \lozenge \textit{charging}$.
In addition, the accepting executions staying in $q_0$ receive the labels \{charging\} or \{charging,working\}, thus satisfying $\lozenge \square charging$.
\end{example}

\subsection{Problem Statement}
We can now formalize the problem considered in this work.

\begin{problem} \label{problem}
For a given LTL task specification $\varphi$ and an SG $\mathcal{G}$ where the transition probabilities are \emph{completely unknown}, design a model-free RL approach to learn an optimal control strategy under which the given LTL tasks are performed successfully with the highest probability against an optimal adversary  (i.e., in the worst case). \update{If the adversary is not optimal, the controller strategy learned for an optimal adversary performs the task with an even larger probability.}
\end{problem}

Formally, our objective is to learn a controller strategy $\mu_\varphi$ for the %
SG $\mathcal{G}$ such that under $\mu_\varphi$, the probability that a path~$\pi$ satisfies the LTL specification $\varphi$ %
is maximized in the worst~case %
\begin{align}
    \mu_\varphi \coloneqq \textnormal{arg}\max_\mu \min_\nu Pr_{\pi \sim \mathcal{G}_{\mu,\nu}}\left\{\pi \mid \pi \models \varphi \right\} \label{eq:ltl_obj}
\end{align}
\update{where $\mu$ and $\nu$ are finite-memory controller and adversarial strategies from Definition \ref{def:strategies}, and $Pr_{\pi \sim \mathcal{G}_{\mu,\nu}}\left\{\pi \mid \pi \models \varphi \right\}$ denotes the probability that a path drawn from $\mathcal{G}_{\mu,\nu}$ satisfies $\varphi$.}

Model-free RL algorithms such as minimax-Q \cite{littman1994} require \update{a discount factor, which can be considered as the probability of continuing the game}, and  a reward function providing a scalar reward after each transition for guidance, and therefore cannot learn directly from the LTL specifications.
Hence, we solve Problem~\ref{problem} by crafting rewards and discount factors from the given LTL specification in a way that a model-free RL algorithm can efficiently learn an optimal strategy from \eqref{eq:ltl_obj} by maximizing the minimum expected value of the return, which is the sum of the discounted rewards.

\section{Reduction from LTL Specifications to Reward Returns} \label{sec:reduction}
In this section, we introduce a reduction from Problem~\ref{problem} to the problem of learning a controller strategy maximizing the worst-case return.
We start by constructing a product game by composing the given SG with the DPA derived from the given LTL specification of the desired task.
We then introduce \emph{novel rewards and discount factors} for each set in the acceptance condition of the DPA, and show that for any memoryless strategy pair, the expected sum of the discounted rewards approaches the probability of satisfying the parity condition as rewards approach zero.

\subsection{Product Game Construction}
By constructing a product game, the problem of learning an optimal control strategy for an LTL task $\varphi$ in an SG~$\mathcal{G}$~is reduced to meeting the parity condition of the derived DPA~$\mathcal{A}_\varphi$. %

\begin{definition}
A \textbf{product game} of an SG $\mathcal{G}$ and a DPA $\mathcal{A}_\varphi$ is a tuple $\mathcal{G}^\times{=}(S^\times, (S_\mu^\times, S_\nu^\times), s_0^\times,\allowbreak A^\times, P^\times, \kappa, C^\times)$ where
$S^\times{=}S{\times} Q$ is the set of product states;  $S_\mu^\times{=}S_\mu{\times} Q$ and $S_\nu^\times{=}S_\nu{\times} Q$ are the controller and adversary product states respectively;
$s_0^\times{=}\langle s_0,q_0\rangle$ is the initial state;
$A^\times{=}A$ with $A^\times(\langle s,q \rangle){=}A(s)$ for all $s{\in} S, q {\in} Q$;
$P^\times{:}S^\times{\times} A^\times{\times} S^\times{\mapsto} [0,1]$ is the probabilistic transition function such that
\begin{align}
    \hspace{-0.8em}P(\langle s,q \rangle, a ,\langle s',q' \rangle){=}\begin{cases}
       P(s,a,s') & \textnormal{if } q'{=}\delta(q,L(s)), \\
       0 & \textnormal{otherwise;}
    \end{cases}
\end{align}
and $C^\times:S^\times\mapsto \{0,\dots,\kappa{-}1\}$ is the product coloring function such that $C^\times(\langle s,q\rangle)=C(q,L(s))$.
A path $\pi^\times=\langle s_0,q_0 \rangle\langle s_1,q_1 \rangle\dots$ in a product game meets the \textbf{parity condition} if it holds that
\begin{align}
    \varphi^\times \coloneqq \textnormal{``}\max\left\{C^\times(s^\times)\mid s^\times\in\textnormal{Inf}^\times(\pi^\times)\right\} \text{ is odd}\textnormal{''},
\end{align}
where $\textnormal{Inf}^\times(\pi^\times)$ is the set of product states visited infinitely many times by $\pi^\times$.
\end{definition}

The product game $\mathcal{G}^\times$ effectively captures this synchronous execution of the SG $\mathcal{G}$ and the DPA $\mathcal{A}_\varphi$; they start in the initial states, and whenever $\mathcal{G}$ moves to a state, $\mathcal{A}_\varphi$ consumes the label of the state and makes a transition.
For example, when $\mathcal{G}$ is in state $s$ and $\mathcal{A}_\varphi$ is in $q$, the product game $\mathcal{G}^\times$ is in $\langle s,q \rangle$. 
If the SG moves to $s'$, the DPA moves to the state $q'{=}\delta(q,L(s'))$, represented by the product game transition~from $\langle s,q \rangle$ to $\langle s',q' \rangle$. %
\update{This implies a one-to-one mapping between the paths of $\mathcal{G}$ and $\mathcal{G}^\times$.}

\begin{lemma} \label{lemma:parity_in_prod_eq_ltl_in_sg}
For a given SG $\mathcal{G}$, a DPA $\mathcal{A}_\varphi$ and their product game $\mathcal{G}^\times$, let $\pi=s_0s_1\dots$ be a path in $\mathcal{G}$ and $\pi^\times=\langle s_0,q_0 \rangle\langle s_1,q_1 \rangle\dots$ be the corresponding path in $\mathcal{G}^\times$. Then, 
\begin{equation}
\left( \pi^\times \models \varphi^\times\right) \Leftrightarrow \left( \pi \models \varphi\right).
\end{equation}
\end{lemma}

\update{
\begin{proof}
It holds from the definition of SGs, as %
the DPA~$\mathcal{A}_\varphi$ makes a transition colored with $k$ if and only if a product state $\langle s,q \rangle$ colored with $k$ is visited in the product game $\mathcal{G}^\times$.
\end{proof}
}

A strategy in $\mathcal{G}^\times$ induces a finite-memory strategy in $\mathcal{G}$ where the states of $\mathcal{A}_\varphi$ act as the modes governed by the transition function of $\mathcal{A}_\varphi$. 
To illustrate this, let $\mu^\times$ denote a \textit{pure memoryless strategy} in $\mathcal{G}^\times$ and $\mu$ denote its \textit{induced strategy} in $\mathcal{G}$. While $\mu$ is operating in mode $m$ corresponding to the DPA state $q$, if a state $s$ is visited in $\mathcal{G}$, $\mu$ changes~its mode from $m$ to $m'$, corresponding to the DPA state $q'{=}\delta(q,L(s'))$; as a result, $\mu$ 
chooses the action that $\mu^\times$ selects in $\langle s',q' \rangle$.

Hence, from Lemma~\ref{lemma:parity_in_prod_eq_ltl_in_sg}, the probability of satisfying the \textit{parity condition} $\varphi^\times$ under a strategy pair $(\mu^
\times,\nu^\times)$ in the product game $\mathcal{G}^
\times$ is equal to the probability of satisfying the LTL specification $\varphi$ in the SG $\mathcal{G}$ under the induced strategy pair $(\mu,\nu)$; this is formalized~in the following lemma.
\begin{lemma} \label{lemma:satisfaction_probabilities_eq_in_prod_and_sg}
For a given strategy pair $(\mu^
\times,\nu^\times)$ in the product game $\mathcal{G}^
\times$ and its induced strategy pair $(\mu,\nu)$ in $\mathcal{G}$, let $\pi$ and $\pi^\times$ be random paths drawn from the MCs $\mathcal{G}_{\mu,\nu}$ and $\mathcal{G}^\times_{\mu^\times,\nu^\times}$ respectively. Then,
\begin{equation}
    Pr\left\{\pi \mid \pi \models \varphi \right\} = Pr\left\{\pi^\times \mid \pi^\times \models \varphi^\times \right\}.
\end{equation}
\end{lemma}
\update{
\begin{proof}
Follows from Lemma~1, since there is a one-to-one mapping  between the paths of $\mathcal{G}_{\mu,\nu}$ and $\mathcal{G}^\times_{\mu^\times,\nu^\times}$.
\end{proof}
}

\begin{example}
Fig.~\ref{fig:prod} presents the product game $\mathcal{G}^\times$ obtained from the SG $\mathcal{G}$ in Fig.~\ref{fig:sg} and the DPA $\mathcal{A}_\varphi$  of the LTL task $\varphi$ in Fig.\ref{fig:dpa}. 
In $\mathcal{G}$, if the adversary follows a pure memoryless strategy, the controller can almost surely win the game in the sense that $\varphi$ is satisfied by taking ``\textnormal{go down}'' in ``\textnormal{Entrance}''. This eventually leads to ``\textnormal{Charger(Agent-On)}'', from which $\mathcal{G}$ can alternate either between ``\textnormal{Charger(Agent-On)}'' and ``\textnormal{Workspace}'' (satisfying $\lozenge \square \textit{working} {\wedge} \square \lozenge \textit{charging}$), 
or between ``\textnormal{Charger(Agent-On)}'' and ``\textnormal{Charger(Agent-Off)}''  (satisfying $\lozenge \square \textit{charging}$); thus, $\varphi$ is satisfied.
However, the adversary can win in the $\mathcal{G}^\times$ by following a pure memoryless strategy that chooses ``\textnormal{turn off}'' in $\langle \textnormal{Charger(Agent-On)},q_0\rangle$ and ``\textnormal{move}'' in $\langle \textnormal{Charger(Agent-On)},q_1\rangle$ as in this case, the maximal color in the obtained infinite cycle 
is even (i.e., 3). This pure memoryless strategy in $\mathcal{G}^\times$ induces a finite-memory strategy in $\mathcal{G}$ that alternates between the modes $m_0$ and $m_1$, under which ``\textnormal{turn off}'' and ``\textnormal{move}'' are selected in ``\textnormal{Charger(Agent-On)}'' respectively, where $m_0$ and $m_1$ correspond to the DPA states $q_0$ and~$q_1$.
\end{example}

We use $Pr_{\mu^\times,\nu^\times}(s^\times \models \varphi^\times)$ to denote the probability that a state $s^\times {\in} S^\times$ satisfies the parity condition $\varphi^\times$ under strategies $(\mu^\times,\nu^\times)$;~i.e.,
\begin{align*}
    Pr_{\mu^\times,\nu^\times}(s^\times \models \varphi^\times) \coloneqq Pr\left\{\pi^\times {\in} \Pi^{\times s^\times}_{\mu^\times,\nu^\times} \mid \pi^\times \models \varphi^\times \right\};
\end{align*}
here, $\Pi^{\times s^\times}_{\mu^\times,\nu^\times}$ denotes the set of all paths of the product MC $\mathcal{G}^{\times,s^\times}_{\mu^\times,\nu^\times}$ that is obtained from $\mathcal{G}^\times_{\mu^\times,\nu^\times}$ by assigning $s^\times$ as the initial state.
Therefore, from Lemma \ref{lemma:satisfaction_probabilities_eq_in_prod_and_sg} and the fact that pure memoryless strategies suffice for the parity condition \cite{chatterjee2012}, our objective can be revised as learning a pure memoryless strategy defined as
\begin{align}
    \mu^\times_{\varphi^\times} \coloneqq \textnormal{arg}\max_{\mu^\times} \min_{\nu^\times} Pr_{\mu^\times,\nu^\times}(s^\times_0 \models \varphi^\times)\label{eq:objective} 
\end{align}
in the product game $\mathcal{G}^\times$. The strategy $\mu^\times_{\varphi^\times}$ is then used to induce the finite-memory strategy $\mu_{\varphi}$ from~\eqref{eq:ltl_obj} in the initial~game~$\mathcal{G}$.

\subsection{Reduction from Parity to Return}
To obtain an optimal strategy defined in \eqref{eq:objective} using model-free RL, we introduce a reward function $R_\varepsilon^\times:S^\times\mapsto\mathbb{R}$ and a \textit{\textbf{state-dependent}} discount function $\gamma_\varepsilon ^\times:S^\times\mapsto[0,1)$, parameterized by $\varepsilon$ as
\begin{align}
    R_\varepsilon^\times(s^\times) &\coloneqq \begin{cases}
       \varepsilon^{\kappa-C^\times(s^\times)} & \textnormal{if } C^\times(s^\times) \textnormal{ is odd}, \\
       0 & \textnormal{if } C^\times(s^\times) \textnormal{ is even},
    \end{cases} \label{eq:rewards} \\
    \gamma_\varepsilon^\times(s^\times) &\coloneqq 1-\varepsilon^{\kappa-C^\times(s^\times)}, \label{eq:discount_factors}
\end{align}
where $\kappa$ is the number of colors.
The idea behind the rewards structure 
is to encourage the agent to repeatedly visit a state colored with an odd number as large as possible by assigning a larger reward to the states with larger odd colors. Further, the rewards are discounted more in the states with larger colors to reflect the parity condition and to keep the return bounded. \update{The parameter $\epsilon$ should be sufficiently small so that the effect of finitely visited states on the return becomes negligible. }

We slightly extend the definition of the return of a suffix of a path $\pi^\times[t{:}]$, denoted by $G^\times_\varepsilon(\pi^\times[t{:}])$, for state-dependent discount factors as
\begin{align}
    G_\varepsilon^\times(\pi^\times[t{:}]) &\coloneqq \sum_{i=t}^\infty \left(\prod_{j=0}^{i-1} \gamma_\varepsilon^\times(\pi^\times[j])\right) R_\varepsilon^\times(\pi^\times[i]).
\end{align}
Now, the return of the entire path is simply $G^\times_\varepsilon(\pi^\times)\coloneqq G^\times_\varepsilon(\pi^\times[0{:}])$. We further use $v^{\times}_{\varepsilon,\mu^\times,\nu^\times}(s^\times)$ to denote the expected return of the paths starting from a state $s^\times$ under a strategy pair $(\mu^\times,\nu^\times)$, which is formally defined as \begin{align}
    v^{\times}_{\varepsilon,\mu^\times,\nu^\times}(s^\times) &\coloneqq \mathbb{E}_{\pi^\times_{s^\times} \sim \mathcal{G}^{\times,s^\times}_{\mu^\times,\nu^\times}} \left[ G_\varepsilon^\times(\pi^\times_{s^\times}) \right].
\end{align}

We \update{establish a connection similar to the Blackwell optimality \cite{hordijk2002} between the values and satisfaction probabilities, and} show that as $\varepsilon$ approaches $0$, the value of each state approaches the %
probability of satisfying $\varphi$ starting from that state.
\begin{theorem} \label{theorem:multiple_discounts_and_rewards}
For a given product game $\mathcal{G}^\times$ and a pure memoryless strategy pair $(\mu^\times,\nu^\times)$, it holds that
\begin{align}
    \lim_{\varepsilon\to 0^+}v^\times_{\varepsilon,\mu^\times,\nu^\times}(s^\times) = Pr_{\mu^\times,\nu^\times}(s^\times \models \varphi^\times)
\end{align}
for any $s^\times\in S^\times$.
\end{theorem}

Before proving Theorem~\ref{theorem:multiple_discounts_and_rewards},  we provide several lemmas. %
Here, $(\mu^\times,\nu^\times)$ denotes an arbitrary but fixed strategy pair in a given product game $\mathcal{G}^\times$; %
we omit the subscripts from $\mathbb{E}_{\pi^\times_{s^\times} \sim \mathcal{G}^{\times,s^\times}_{\mu^\times,\nu^\times}}$, $Pr_{\mu^\times,\nu^\times}$, $v^{\times}_{\varepsilon,\mu^\times,\nu^\times}$ and use $\mathbb{E}$, $Pr$, $v^\times_\varepsilon$ instead. 

We first establish bounds on the returns of the paths of $\mathcal{G}^\times$. In the following lemma, the first inequality~\eqref{eq:g_is_between_0_and_1} states that the returns and thereby the values are always between 0 and 1. The second inequality~\eqref{eq:changing_to_larger_odd_colors_increases_returns} states that changing the color of a state on a path to a larger odd number increases the return of the path. Similarly, the third inequality~\eqref{eq:changing_to_larger_even_colors_decreases_returns} states that changing the color to a larger even number decreases the return.

\begin{lemma} \label{lemma:bounds_on_returns_in_product_games}
For a given product game $\mathcal{G}^\times$, the following hold for any path $\pi^\times$ and any time step $t\geq0$:
\begin{align}
    &0 \leq G_\varepsilon^\times(\pi^\times[t{:}]) \leq 1, \label{eq:g_is_between_0_and_1} \\
    &G_\varepsilon^\times(\pi^\times[t{:}]) \leq \varepsilon^{\kappa-k}{+}(1{-}\varepsilon^{\kappa-k})G_\varepsilon^\times(\pi^\times[t{+}1{:}]), \label{eq:changing_to_larger_odd_colors_increases_returns}\\
    &G_\varepsilon^\times(\pi^\times[t{:}]) \geq (1{-}\varepsilon^{\kappa-k})G_\varepsilon^\times(\pi^\times[t{+}1{:}]), \label{eq:changing_to_larger_even_colors_decreases_returns}
\end{align}
where $k=\min\{\kappa,C^\times(\pi^\times[t])+1\}$.
\end{lemma}

\begin{proof}
The lower bound in \eqref{eq:g_is_between_0_and_1} holds since there is no negative reward. Now, assume that we can change the colors of the states at each time step. In this case, the maximum return can be obtained by assigning the same odd color to all the states, which results in a return $\varepsilon^{\kappa-k} + \varepsilon^{\kappa-k}(1-\varepsilon^{\kappa-k}) + \varepsilon^{\kappa-k}(1-\varepsilon^{\kappa-k})^2+ \dots = 1$, concluding the upper bound in \eqref{eq:g_is_between_0_and_1} holds. By definition, for a given color $k'$, the return can 

\noindent be written recursively as
\begin{align}
    \hspace{-2pt}G_\varepsilon^\times(\pi^\times[t{:}]) &{=} \begin{cases}
       \varepsilon^{\kappa-k'} {+} (1{-}\varepsilon^{\kappa-k'})G_\varepsilon^\times(\pi^\times[t{+}1{:}]) & \textnormal{if } k' \textnormal{ is odd}, \\
       (1{-}\varepsilon^{\kappa-k'})G_\varepsilon^\times(\pi^\times[t{+}1{:}]) & \textnormal{if } k' \textnormal{ is even}.
    \end{cases}\hspace{-2pt} \label{eq:return_recursive}
\end{align}
From \eqref{eq:return_recursive}, \eqref{eq:g_is_between_0_and_1}, and $\varepsilon\in(0,1]$, we obtain \eqref{eq:changing_to_larger_odd_colors_increases_returns} and \eqref{eq:changing_to_larger_even_colors_decreases_returns} as:
\begin{align}
    G_\varepsilon^\times(\pi^\times[t{:}])&\leq \varepsilon^{\kappa-k'} + (1{-}\varepsilon^{\kappa-k'})G_\varepsilon^\times(\pi^\times[t{+}1{:}]) \notag \\
    &=\varepsilon^{\kappa-k'}(1{-}G_\varepsilon^\times(\pi^\times[t{+}1{:}]))+G_\varepsilon^\times(\pi^\times[t{+}1{:}]) \notag \\ 
    &\leq\varepsilon^{\kappa-k}(1{-}G_\varepsilon^\times(\pi^\times[t{+}1{:}]))+G_\varepsilon^\times(\pi^\times[t{+}1{:}]) \notag \\ 
    &= \varepsilon^{\kappa-k}+(1{-}\varepsilon^{\kappa-k})G_\varepsilon^\times(\pi^\times[t{+}1{:}]), \\
    G_\varepsilon^\times(\pi^\times[t{:}])&\geq (1{-}\varepsilon^{\kappa-k'})G_\varepsilon^\times(\pi^\times[t{+}1{:}]) \notag \\
    &\geq (1{-}\varepsilon^{\kappa-k})G_\varepsilon^\times(\pi^\times[t{+}1{:}]),
\end{align}
since $k'\leq k$.
\end{proof}

The parity condition of a product game is defined over the recurrent states, i.e., the states that are visited infinitely many times, in a product game. In the MC induced by a strategy pair, a state is recurrent if and only if it belongs to a BSCC of the MC as the paths almost surely reach a BSCC and visit its states infinitely often.
If the largest color among the colors of the states in a BSCC is an odd number, the BSCC is called \textbf{\emph{accepting}} as all the paths reaching the BSCC satisfy the parity condition. Similarly, if the largest color is an even number, then the BSCC is called \textbf{\emph{rejecting}}. 
Thus, the probability of satisfying the parity condition can be obtained by computing the probability of reaching an accepting BSCC (see the proofs provided for the Rabin condition in \cite{baier2008}). 

We now show that the values of the states in a BSCC approach  $1$ if the BSCC is accepting and 0 otherwise. Using this lemma, we then prove Theorem \ref{theorem:multiple_discounts_and_rewards} by showing that the values approach the probability of reaching accepting BSCCs.

\begin{lemma} \label{lemma:values_in_bsccs_are_1_if_accepting_0_otherwise}
For a BSCC $\mathcal{V}^\times\in\mathcal{B}(\mathcal{G}^\times_{\mu^\times,\nu^\times})$, and for all $s^\times \in \mathcal{V}^\times$, it holds that
\vspace{-2pt}
\begin{align}
    \lim_{\varepsilon\to 0^+}v^{\times}_{\varepsilon,\mu^\times,\nu^\times}(s^\times) = \begin{cases}
    1 & \textnormal{if } \mathcal{V}^\times \textnormal{ is accepting}, \\
    0 & \textnormal{if } \mathcal{V}^\times \textnormal{ is rejecting}.
    \end{cases} \label{eq:values_in_bsccs_are_1_if_accepting_0_otherwise}
\end{align}
\end{lemma}

\begin{proof}
To simplify our notation, for a BSCC $\mathcal{V}^\times\in\mathcal{B}(\mathcal{G}^\times_{\mu^\times,\nu^\times})$, we let $k_{\mathcal{V}^\times}$ denote the largest color among the colors of the states in $\mathcal{V}^\times$; i.e.,
\vspace{-2pt}
\begin{align}
    k_{\mathcal{V}^\times}\coloneqq\max\left\{C^\times(s^\times)\mid s^\times\in\mathcal{V}^\times\right\},
\end{align}
and we let $r_{\mathcal{V}^\times}$ denote $\varepsilon^{\kappa-k_{\mathcal{V}^\times}}$.

We first consider the case where $k_{\mathcal{V}^\times}$ is an odd number; i.e., $\mathcal{V}^\times$ is accepting. Let $s^\times_* \in \mathcal{V}^\times$ be a state colored with $k_{\mathcal{V}^\times}$.
As $\mathcal{V}^\times$ is a BSCC, a path $\pi^\times_{s^\times_*}$, starting from $s^\times_*$, almost surely visits $s^\times_*$ and the other states in $\mathcal{V}^\times$ infinitely many times. By Lemma~\ref{lemma:bounds_on_returns_in_product_games}, we can obtain a lower bound on $v^{\times}_{\varepsilon}(s^\times_*)$, the expected value of the return of $\pi^\times_{s^\times_*}$, by replacing the colors of the states other than $s^\times_*$ by $k_{\mathcal{V}^\times}{-}1$.
Let $N$ be the number of time steps between two consecutive visits to $s^\times_*$; then, %
\begin{align*}
    \hspace{-2px}v^{\times}_{\varepsilon}(s^\times_*) \geq r_{\mathcal{V}^\times}{+}(1{-}r_{\mathcal{V}^\times})\mathbb{E} \left[ (1{-}\varepsilon r_{\mathcal{V}^\times})^N G_\varepsilon^\times(\pi^\times_{s^\times_*}[N{:}]) \right]. \hspace{-2px}
\end{align*}
Now, by Jensen's inequality and the Markov property, we can simplify the bound as
\begin{align}
\label{eq:temp1}
    v^{\times}_{\varepsilon}(s^\times_*)\geq r_{\mathcal{V}^\times}{+}(1{-}r_{\mathcal{V}^\times}) (1{-}\varepsilon r_{\mathcal{V}^\times})^n v^{\times}_{\varepsilon}(s^\times_*),
\end{align}
where $n\coloneqq \mathbb{E}[N]$. From %
\eqref{eq:temp1} and the fact that $v^{\times}_{\varepsilon}(s^\times_*)$, $r_{\mathcal{V}^\times}$, and $\varepsilon$ are all between 0 and 1, we further obtain that
\begin{align}
    v^{\times}_{\varepsilon}(s^\times_*) &\geq \frac{r_{\mathcal{V}^\times}}{1-(1{-}r_{\mathcal{V}^\times}) (1{-}\varepsilon r_{\mathcal{V}^\times})^{n}} \notag \\
    &\geq \frac{r_{\mathcal{V}^\times}}{1-(1{-}r_{\mathcal{V}^\times}) (1{-}n\varepsilon r_{\mathcal{V}^\times})} \notag \\
    &= \frac{1}{1-n\varepsilon(1-r_{\mathcal{V}^\times})}.
\end{align}
As $\varepsilon$ goes to $0$, $n\varepsilon(1-r_{\mathcal{V}^\times})$ goes to $0$ as well, which makes $v^{\times}_{\varepsilon}(s^\times_*)$ go to 1. 

Furthermore, we can derive a lower bound on the value of any state $s^\times\in\mathcal{V}^\times$ in a similar way as
\begin{align}
    v^{\times}_{\varepsilon}(s^\times) \geq (1{-}\varepsilon r_{\mathcal{V}^\times})^{n'} v^{\times}_{\varepsilon}(s^\times_*)
\end{align}
where $n'$ is the expected number of time steps between leaving $s^\times$ and reaching $s^\times_*$. Since $n'$ is a constant, $(1{-}\varepsilon r_{\mathcal{V}^\times})^{n'}$ along with $v^{\times}_{\varepsilon}(s^\times_*)$ goes to $1$ as $\varepsilon$ goes to $0$, making $v^{\times}_{\varepsilon}(s^\times)$ goes to $1$. Thus, since the lower bounds go to 1 when $\varepsilon \to 0^+$ and the values are bounded above by $1$ due to \eqref{eq:g_is_between_0_and_1}, we have that~\eqref{eq:values_in_bsccs_are_1_if_accepting_0_otherwise} holds when $\mathcal{V}^\times$ is accepting.

We now consider the case where $k_{\mathcal{V}^\times}$ is an even number (i.e., $\mathcal{V}^\times$ is rejecting). Similarly to the previous case, from Lemma~\ref{lemma:bounds_on_returns_in_product_games}, we obtain the following upper bound by changing the colors of the states other than $s^\times_*$ with $k_{\mathcal{V}^\times}{-}1$ -- i.e.,
\begin{align}
    v^{\times}_{\varepsilon}(s^\times_*) \leq (1{-}r_{\mathcal{V}^\times})\mathbb{E} \Big[&1-(1{-}\varepsilon r_{\mathcal{V}^\times})^N \notag \\
    &+(1{-}\varepsilon r_{\mathcal{V}^\times})^N G_\varepsilon^\times(\pi^\times_{s^\times_*}[N{:}]) \Big],
\end{align}
where $N$ is defined as in the previous case. Again, from Jensen's inequality, the Markov property, and the fact that $1{-}\varepsilon r_{\mathcal{V}^\times}$ and $ 1{-}r_{\mathcal{V}^\times}$ are between 0 and 1, we obtain that
\begin{align}
    v^{\times}_{\varepsilon}(s^\times_*) \leq 1{-}(1{-}\varepsilon r_{\mathcal{V}^\times})^n {+} (1{-}r_{\mathcal{V}^\times}) v^{\times}_{\varepsilon}(s^\times_*),
\end{align}
where $n'$ is expected value of $N$. Hence, it holds that
\begin{align}
\label{eq:temp2}
    v^{\times}_{\varepsilon}(s^\times_*) \leq \frac{1{-}(1{-}\varepsilon r_{\mathcal{V}^\times})^n}{r_{\mathcal{V}^\times}} \leq \frac{n\varepsilon r_{\mathcal{V}^\times}}{r_{\mathcal{V}^\times}} = n\varepsilon.
\end{align}
Using %
\eqref{eq:temp2}, we obtain an upper bound on any state $s^\times\in\mathcal{V}^\times$~as
\begin{align*}
    v^{\times}_{\varepsilon}(s^\times) \leq (1{-}\varepsilon r_{\mathcal{V}^\times})^{n'}+v^{\times}_{\varepsilon}(s^\times_*) \leq (1{-}\varepsilon r_{\mathcal{V}^\times})^{n'}+n\varepsilon,
\end{align*}
where $n'$ is defined as in the previous case. Since $n$ and $n'$ are constant, this upper bound goes to $0$ as $\varepsilon$ goes to $0$, which, by \eqref{eq:g_is_between_0_and_1}, concludes the proof.
\end{proof}

With the above results, we now prove Theorem~\ref{theorem:multiple_discounts_and_rewards}.

\begin{proof}[Theorem~\ref{theorem:multiple_discounts_and_rewards}]
The probability that the parity condition is satisfied from a state $s^\times\in S^\times$, denoted by $Pr(s^\times \models \varphi^\times)$, is equivalent to the probability that a path $\pi^\times_{s^\times}$, which starts in $s^\times$, reaches an accepting BSCC.
Let $\mathcal{U}^\times_+$ and $\mathcal{U}^\times_-$ denote the union of all the states belonging to accepting and rejecting BSCCs respectively; then we can express the value~as
\begin{align}
    v^{\times}_{\varepsilon}(s^\times) &= \mathbb{E}\left[G^\times_\varepsilon(\pi^\times_{s^\times})\mid \pi^\times_{s^\times} \models \lozenge \mathcal{U}^\times_+\right] Pr(s^\times \models \varphi^\times) \notag \\
    &+ \mathbb{E}\left[G^\times_\varepsilon(\pi^\times_{s^\times})\mid \pi^\times_{s^\times} \models \lozenge \mathcal{U}^\times_- \right] Pr(s^\times \not\models \varphi^\times).
\end{align}

Let $m_+$ denote the expected number of time steps until an accepting BSCC is reached. Using Lemma~\ref{lemma:bounds_on_returns_in_product_games}, we can obtain a lower bound on the value by changing the colors of all transient states to the largest even number; i.e.,
\begin{align}
    v^\times_\varepsilon(s^\times) \geq (1{-}\varepsilon)^{m_+} \underline{v^\times_\varepsilon}(\mathcal{U}^\times_+)Pr(s^\times \models \varphi^\times), \label{eq:v_lower_bound}
\end{align}
where $\underline{v^\times_\varepsilon}(\mathcal{U}^\times_+) \coloneqq \min_{s^\times \in \mathcal{U}^\times_+} v^\times_\varepsilon(s^\times)$. 
From Lemma~\ref{lemma:values_in_bsccs_are_1_if_accepting_0_otherwise}, this lower bound goes to $Pr(s^\times \models \varphi^\times)$ as $\varepsilon$ goes to $0$.

Similarly, let $m_-$ denote the expected number of time steps until a rejecting BSCC is reached. As changing the colors of all transient states to the largest odd number increases the return due to Lemma~\ref{lemma:bounds_on_returns_in_product_games}, the following upper bound on the value
\begin{align}
    v^\times_\varepsilon(s^\times) &\leq (1{-}\varepsilon r_{\mathcal{V}^\times})^{m_-} \overline{v^\times_\varepsilon}(\mathcal{U}^\times_-)  Pr(s^\times \not\models \varphi^\times) \notag \\
    & \hspace{90px} +Pr(s^\times \models \varphi^\times) \label{eq:v_upper_bound}
\end{align}
holds, where $\overline{v^\times_\varepsilon}(\mathcal{U}^\times_-) \coloneqq \max_{s^\times \in \mathcal{U}^\times_-} v^\times_\varepsilon(s^\times)$; this bound goes to $0$ as $\varepsilon$ goes to $0$ due to Lemma \ref{lemma:values_in_bsccs_are_1_if_accepting_0_otherwise}. 

Since both the lower bound~\eqref{eq:v_lower_bound} and the upper bound~\eqref{eq:v_upper_bound} go to $Pr(s^\times \models \varphi^\times)$ as $\varepsilon\to 0^+$, it follows that $v^\times_\varepsilon(s^\times)$ goes to $Pr(s^\times \models \varphi^\times)$ as $\varepsilon\to0^+$, concluding the proof.
\end{proof}

\section{Lazy Color Generation} 
\label{sec:lazy}

In this section, we introduce a framework that efficiently uses the rewards and discount factors established in the previous section to learn optimal strategies. 
For a given number of colors $\kappa$, the functions in \eqref{eq:rewards} and \eqref{eq:discount_factors} provide $\kappa$ distinct rewards and discount factors. 
Larger discount factors are used in the states with smaller colors so that visiting such states affects the return less than the ones with larger colors. 
However, if $\kappa$ is large, the discount factors could be too large for the smaller colors.
For example, for $\varepsilon=0.01$ and $\kappa=7$, the discount factor for the color $0$ is very close to~1 (i.e., $1{-}10^{-14}$) and thus would extremely slowdown learning~convergence. 

To address this problem, we modify the design of the product game; %
we start with a product game having only a single color and increase the number of colors only when a new color is needed. The key idea is that if the large colors are not necessary to learn an optimal strategy, then there is no need to specify distinct reward and discount factors for them; thus, smaller discount factors can be also used for small~colors.

\begin{definition}
A \textbf{multilevel product game} (MPG) $\mathcal{G}^\star$ of a product game $\mathcal{G}^\times$ is a tuple $\mathcal{G}^\star{=}(\kappa, S^\star, (S_\mu^\star, S_\nu^\star), s_0^\star,\allowbreak A^\star, P^\star, R^\star, \gamma^\star)$ where
$\kappa$ is the number of colors (levels) and $[\kappa]\coloneqq\{1,\dots,\kappa\}$;
$S^\star{=}S^\times{\times}[\kappa]$ is the set of multilevel product states;  $S_\mu^\star{=}S_\mu^\times{\times}[\kappa]$ and $S_\nu^\star{=}S_\nu^\times{\times}[\kappa]$ are the controller and adversary multilevel product states respectively;
$s_0^\star{=}\langle s_0^\times,1\rangle$ is the initial state;
$A^\star{=}A^\times$ is the set of actions;
$P^\star{:}S^\star{\times} A^\star{\times} S^\star{\mapsto} \allowbreak[0,1]$ is the probabilistic transition function such that $P^\star(\langle s^\times, \bar{\kappa} \rangle, a^\star, \langle {s^\times}', \bar{\kappa}' \rangle ) := $
\begin{align}
\label{eq:temp3}
    P^\times (s^\times, a^\star, {s^\times}')\cdot\begin{cases}
    \tau_{\varepsilon} & \textnormal{if } \bar{\kappa}'>\bar{\kappa} \textnormal{ and } C^\times(s^\times)=\bar{\kappa}'{-}1, \\
    1{-}\tau_{\varepsilon} & \textnormal{if } \bar{\kappa}'=\bar{\kappa} \textnormal{ and } C^\times(s^\times)\geq\bar{\kappa}, \\
    1 & \textnormal{if } \bar{\kappa}'=\bar{\kappa} \textnormal{ and } C^\times(s^\times)<\bar{\kappa}, \\
    0 & \textnormal{otherwise};
    \end{cases}
\end{align}
where $\tau_\varepsilon$ is a function of $\varepsilon$ satisfying $\lim_{\varepsilon \to 0^+} \frac{\varepsilon}{\tau_\varepsilon} = 0$ and $R^\star:S^\star\mapsto[0,1]$ and $\gamma^\star:S^\star\mapsto[0,1]$ are the reward and discount functions respectively such that 
\begin{align}
    R_\varepsilon^\star(\langle s^\times,\bar{\kappa}\rangle) &\coloneqq \begin{cases}
       \varepsilon^{\bar{\kappa}-C^\star(\langle s^\times,\bar{\kappa} \rangle)} & \textnormal{if } C^\star(\langle s^\times,\bar{\kappa} \rangle) \textnormal{ is odd}, \\
       0 & \textnormal{if } C^\star(\langle s^\times,\bar{\kappa} \rangle) \textnormal{ is even},
    \end{cases} \label{eq:rewards_reformulated} \\
    \gamma_\varepsilon^\star(\langle s^\times,\bar{\kappa} \rangle) &\coloneqq 1-\varepsilon^{\bar{\kappa}-C^\star(\langle s^\times,\bar{\kappa}\rangle)}, \label{eq:discount_factors_reformulated}
\end{align}
where $C^\star(\langle s^\times,\bar{\kappa}\rangle ) \coloneqq \min\{C^\times(s^\times),\bar{\kappa}{-}1\}$.
\end{definition}

The MPG $\mathcal{G}^\star$ of a product game $\mathcal{G}^\times$ has a copy of $\mathcal{G}^\times$ for each level $\bar{\kappa}\in \{1,\dots,\kappa\}$.
In the $\bar{\kappa}$-th level, the reward and discount factors are specified as if $\bar{\kappa}$ is the total number of colors and $\bar{\kappa}{-}1$ is the largest color; thus, the colors greater than $\bar{\kappa}{-}1$ {in the $\bar{\kappa}$-th level copy of $\mathcal{G}^\times$ are} truncated to $\bar{\kappa}{-}1$.
This results in smaller discount factors and larger rewards for lower levels, thereby speeding up learning.

In addition, as long as the colors of the states visited infinitely often are less than $\bar{\kappa}$ in the $\bar{\kappa}$-th level, the return of a path approaches the probability of satisfying the parity condition as $\varepsilon$ goes to $0$. This is because the effect of the states visited only for a finite number of times on the return vanishes as $\varepsilon$ goes to $0$, as shown in the proof of Theorem~\ref{theorem:multiple_discounts_and_rewards}. 
Therefore, truncating the larger colors to $\bar{\kappa}{-}1$ does not change the return in the limit. 
However, this does not hold if a state with a color $k\geq\bar{\kappa}$ is visited infinitely often as it might alter the satisfaction of the parity condition.
In such cases, a transition to the $(k{+}1)$-th level needs to be made to reflect the parity~condition.

The main challenge here is that since the transition model of the initial SG $\mathcal{G}$ (and thus $\mathcal{G}^\times$ and $\mathcal{G}^\star$) is \emph{unknown}, it is not possible to determine in advance whether a state with a color $k\geq\bar{\kappa}$ is visited infinitely often under a strategy pair. Consequently, our key idea is to allow for probabilistic transitions between levels so that repeated visits can lead to upper levels.
Specifically, the transitions of the multilevel game $\mathcal{G}^\star$ are constructed in~\eqref{eq:temp3} in a way that it is impossible to make a transition from the $\bar{\kappa}$-th level to the $\bar{\kappa}'$-th level if $\bar{\kappa}'<\bar{\kappa}$, and a transition from a state $s^\times$ in the $\bar{\kappa}$-th level to the $\bar{\kappa}'$-th level can happen w.p. $\tau_{\varepsilon}$ (i.e., stays in the same level w.p. $1-\tau_{\varepsilon}$) only if $\bar{\kappa}'>\bar{\kappa}$ and $C^\times(s^\times)=\bar{\kappa}'$.
Such probabilistic transitions ensure that repeatedly visiting a state with a color $k\geq\bar{\kappa}$  in the $\bar{\kappa}$-th level almost surely results in a transition to the level where the color $k$ is the largest~color. 

This captures our goal that, in order to accelerate learning, \emph{the larger colors in the parity condition should be considered only when needed}. If the first $\bar{\kappa}$ colors are sufficient to reflect the parity condition for the strategies being followed, $\mathcal{G}^\star$ stays in the $\bar{\kappa}$-th level and uses only $\bar{\kappa}$ distinct rewards and discount factors; and if they are not, $\mathcal{G}^\star$ eventually moves to an upper level where more colors are considered to provide a sufficient number of distinct rewards and discount factors. However, the transitions to the lower levels are not allowed to avoid redundant circular transitions between the levels as they can possibly distort the returns.

Now, any pure memoryless strategy pair $(\mu^\star,\nu^\star)$ in the MPG $\mathcal{G}^\star$ of a product game $\mathcal{G}^\times$ induces a finite-memory strategy pair $(\mu^\times,\nu^\times)$ in $\mathcal{G}^\times$, where the levels of $\mathcal{G}^\star$ act as modes; i.e., whenever $\mathcal{G}^\star$ transitions to an upper level, the strategies $\mu^\times$ and $\nu^\times$ switch to the corresponding mode. %
Hence, we below capture the result that enables the use of MPGs; we show that under a strategy pair, the values of the states in $\mathcal{G}^\star$ approach the probability of satisfying the parity condition in $\mathcal{G}^\times$ as $\varepsilon$ goes~to~0.

\begin{theorem}\label{theorem:multilevel_values}
For the MPG $\mathcal{G}^\star$ of a product game $\mathcal{G}^\times$, let $(\mu^\star,\nu^\star)$ be an arbitrary pure memoryless strategy pair in $\mathcal{G}^\star$ and let $(\mu^\times,\nu^\times)$ be its induced strategy pair in $\mathcal{G}^\times$. Then, it holds that
\begin{align}
    \lim_{\varepsilon\to 0^+}v^\star_{\varepsilon,\mu^\star,\nu^\star}(\langle s^\times, \bar{\kappa}\rangle) = Pr_{\mu^\times,\nu^\times}(s^\times \models \varphi^\times)
\end{align}
for any $s^\times\in S^\times$ and $\bar{\kappa}\in \{1,\dots,\kappa\}$.
\end{theorem}

\begin{proof}
A BSCC of the induced MC $\mathcal{G}^\star_{\mu^\star,\nu^\star}$ cannot contain states belonging to the different levels since transitions from an upper level to a lower level are not permitted. Moreover, a BSCC in the $\bar{\kappa}$-th level cannot have a state whose color is greater than or equal to $\bar{\kappa}$ because otherwise a transition to an upper level eventually happens, conflicting with the BSSC definition.
This means that none of the colors in a BSCC is truncated, i.e., there is a distinct reward and a discount factor for each color. Thus, from Lemma~\ref{lemma:values_in_bsccs_are_1_if_accepting_0_otherwise}, all the values in a BSCC approach either $1$ or $0$ depending on whether the BSCC is accepting or rejecting. 

We can now  complete the proof by following the same steps as in Theorem \ref{theorem:multiple_discounts_and_rewards} proof. The only difference in this case is that the number of steps before reaching a BSCC might depend on the parameter $\varepsilon$ and $\kappa$ in $\mathcal{G}^\star_{\mu^\star,\nu^\star}$. However, since $\tau_{\varepsilon}$ approaches 0 slower than $\varepsilon$, both $(1{-}\varepsilon)^{\kappa m_+/\tau_{\varepsilon}}$ and $(1{-}\varepsilon r_{\mathcal{V}^\times})^{\kappa m_-/\tau_{\varepsilon}}$ go to $1$ as $\varepsilon$ goes to 0, the same results hold.
\end{proof}

In the MPGs, the probability of transitioning to an upper level is controlled by the parameter $\tau_{\varepsilon}$ (see~\eqref{eq:temp3}). A smaller $\tau_{\varepsilon}$ encourages the controller and the adversary
to learn optimal strategies utilizing a fewer number of colors. For many SGs for which such optimal strategies exist, this can result in faster convergence since the discount factors to be used are smaller, as we demonstrate in our case studies.
In addition, the space complexity of the MPGs is \emph{only linear} in the total number of colors, and any model-free RL algorithm is guaranteed to converge to an optimal controller strategy maximizing the minimum satisfaction probability for a sufficiently small $\varepsilon$, as %
formalized~below.

\begin{theorem} \label{theorem:lazy_color_generation}
For the MPG $\mathcal{G}^\star$ constructed from the product game $\mathcal{G}^\star$ of a given SG $\mathcal{G}$ and a given LTL specification $\varphi$, there exists $\varepsilon'>0$ such that for any $\varepsilon \in (0,\varepsilon']$, a model-free RL algorithm, such as minimax-Q, converges to a pure memoryless controller strategy $\mu^\star_*$ in $\mathcal{G}^\star$, and its induced strategy $\mu_\varphi$ maximizes the minimum probability of satisfying $\varphi$ in $\mathcal{G}$ -- i.e., $\mu_\varphi$ satisfies~\eqref{eq:ltl_obj}.
\end{theorem}

\begin{proof}
Since there are only finitely many different pure memoryless strategies that can be followed in an MPG, the number of possible different satisfaction probabilities that can be obtained by them is also finite. Thus, the absolute differences between the satisfaction probabilities for different strategy pairs are either $0$ or greater than for some $d>0$. %
Hence, from Theorem~\ref{theorem:multilevel_values}, for the MPG $\mathcal{G}^\star$ of the product game $\mathcal{G}^\times$ constructed from an SG $\mathcal{G}$ and an LTL task $\varphi$, there exists $\varepsilon'>0$ such that for all $\varepsilon \in (0,\varepsilon']$, it holds that
\begin{equation}
\begin{split}
    \hspace{-10pt}v^\star_{\varepsilon,\mu_1^\star,\nu_1^\star}(\langle s^\times, \bar{\kappa}\rangle) &>    v^\star_{\varepsilon,\mu_2^\star,\nu_2^\star}(\langle s^\times, \bar{\kappa}\rangle) \\
    \hspace{-10pt}\iff Pr_{\mu_1^\times,\nu_1^\times}(s^\times \models \varphi^\times) &> Pr_{\mu_2^\times,\nu_2^\times}(s^\times \models \varphi^\times)
\end{split}
\end{equation}
for any two pure memoryless strategy pairs $(\mu_1^\star,\nu_1^\star)$ and $(\mu_2^\star,\nu_2^\star)$ in $\mathcal{G}^\star$ and their induced strategy pairs $(\mu_1^\times,\nu_1^\times)$ and $(\mu_2^\times,\nu_2^\times)$ in $\mathcal{G}^\times$. 

As a result, a memoryless controller strategy $\mu_*^\star$ maximizing the minimum expected return in $\mathcal{G}^\star$ induces a finite-memory strategy $\mu_{\varphi^\times}^\times$ maximizing the minimum probability of satisfying the parity condition in $\mathcal{G}^\times$. 
Additionally, from Lemmas~\ref{lemma:parity_in_prod_eq_ltl_in_sg} and~\ref{lemma:satisfaction_probabilities_eq_in_prod_and_sg}, $\mu_{\varphi^\times}^\times$ further induces a finite-memory strategy $\mu_\varphi$ in $\mathcal{G}$ that maximizes the probability of satisfying $\varphi$ in the worst case.
Finally, since all the discount factors are strictly less than $1$ and all the rewards are Markovian, under regular conditions on the learning rate and the exploration strategy, a model-free RL algorithm such as minimax-Q~\cite{littman1994} is guaranteed to converge to such optimal~strategies.
\update{We note that using state-dependent discount factors does not violate the convergence conditions of minimax-Q. The largest discount factor in \eqref{eq:discount_factors_reformulated},
i.e., $(1{-}\epsilon^{\kappa-1})$, which is less than $1$, can be used as the contraction index in the convergence proofs (e.g., see~\cite{littman1996})}. 

\update{Our result also suggests a PAC algorithm based on the minimum transition probability as discussed in~\cite{yang2022}. However, the convergence rate of such an algorithm could be arbitrarily slow depending on the minimum transition probability in the stochastic game. In cases where multiple optimal controller strategies exist, the converged strategies tend to satisfy the specifications as quickly as possible due to discounting.}
\end{proof}

\begin{algorithm}[!t]
\caption{\small Model-free RL for LTL tasks.}
\label{algorithm}
\begin{algorithmic}
\begin{small}
\REQUIRE{LTL formula $\varphi$, unknown SG $\mathcal{G}$, parameter $\varepsilon$}
\ENSURE{optimal controller strategy $\mu_\varphi$}
\STATE Translate $\varphi$ to a DPA $\mathcal{A}_\varphi$
\STATE Construct a product game $\mathcal{G}^\times$ by composing $\mathcal{G}$ and $\mathcal{A}_\varphi$
\STATE Construct the MPG $\mathcal{G}^\star$ from
$\mathcal{G}^\times$
\STATE 
Initialize $Q(s^\star, a)$ for each state $s^\star$ and action $a$
\FOR [  \# for each episode ]{$i=0$ {\bfseries to} $\mathcal{I}-1$}
\FOR [  \# for each time step ]{$t=0$ {\bfseries to} $\mathcal{T}-1$}
\STATE Derive an $e$-greedy strategy pair $(\mu^\star,\nu^\star)$ from $Q$
\STATE Take the action $a_t \leftarrow \begin{cases}\mu^\star(s^\star_t), & s^\star_t {\in} S^\star_\mu \\ \nu^\star(s^\star_t), & s^\star_t {\in} S^\star_\nu \end{cases}$
\STATE Observe the next state $s^\star_{t+1}$
\STATE $Q(s^\star_t,a_t) \leftarrow (1-\alpha_{_{i,t}}) Q(s^\star_t,a_t) + \alpha_{_{i,t}}  R^\star(s^\star_t)$
\STATE \hspace{40pt} $+ \alpha_{_{i,t}}  \gamma^\star(s^\star_t) \cdot \begin{cases}\max_{a'} Q(s^\star_{t+1},a'), & s^\star_{t+1} {\in} S^\star_\mu \\ \min_{a'} Q(s^\star_{t+1},a'), & s^\star_{t+1} {\in} S^\star_\nu \end{cases}$
\ENDFOR
\ENDFOR
\STATE Derive a greedy controller strategy $\mu^\star_*$ from $Q$
\STATE Obtain the induced controller strategy $\mu_\varphi$ for $\mathcal{G}$ from $\mu^\star_*$ from $Q$
\end{small}
\end{algorithmic}
\end{algorithm}

The overall approach is summarized in Algorithm~\ref{algorithm}, which takes as input an LTL formula $\varphi$, an SG $\mathcal{G}$ and a parameter $\varepsilon$. 
The algorithm first translates $\varphi$ into a DPA $\mathcal{A}_\varphi$ and composes it with $\mathcal{G}$ to construct the product game $\mathcal{G}^\times$. 
Then, it obtains the MPG $\mathcal{G}^\star$ of $\mathcal{G}^\times$. Since $\mathcal{G}$ is unknown, the transition probabilities and the graph topologies of $\mathcal{G}$, $\mathcal{G}^\times$, and $\mathcal{G}^\star$ are~unknown. 

In each episode, Algorithm~\ref{algorithm} starts in the initial state and
follows a $e$-greedy strategy pair, under which a random action is taken w.p. $e$ and a greedy action is taken w.p. $1-e$. At each time step $t$, after taking an action $a$, the next state $s^\star_{t+1}$ is observed and $Q(s^\star_t, a)$ is updated according to the minimax-Q algorithm \cite{littman1994} using a properly decreasing learning rate $\alpha_{_{t,i}}$. Each episode terminates after $\mathcal{T}$ time steps, and after $\mathcal{I}$ episodes, a greedy controller strategy is derived from the Q-values, then used to induce the controller strategy for~$\mathcal{G}$.

\section{Three-Color Approximation} \label{sec:three_color}

In this section, we provide a modified version of the MPGs for the scenarios where the controller can learn optimal strategies by eventually focusing only on a single odd color. The idea here is that controller nondeterministically decides to focus on an odd color $k$, and once the controller decides, the colors smaller than $k$ are replaced with the color $0$, the colors larger than $k$ are replaced with the color $2$, and finally, the color $k$ is replaced with the color $1$. This results in three distinct reward and discount factors regardless of the actual number of colors; thus, improving the convergence rate in the worst case compared to the lazy color generation. 

Such an approach, however, may not yield optimal strategies if the adversary can observe and make use of the controller's decision. Nonetheless, in many scenarios, if the controller carefully chooses when to decide \update{to ensure that, after the decision, the adversary cannot follow a strategy under which a state with a color larger than the focused color is repeatedly visited}, the adversary cannot benefit from observing the decision. 

To reflect this approach, we modify the MPGs as follows.
\begin{definition}
A \textbf{three-color multilevel product game} (\textbf{TMPG}) $\mathcal{G}^{{\hat{\star}}}$ of a product game $\mathcal{G}^\times$ is a tuple $\mathcal{G}^{{\hat{\star}}} = (\kappa, S^{{\hat{\star}}}, \allowbreak (S_\mu^{{\hat{\star}}}, S_\nu^{{\hat{\star}}}),  s_0^{{\hat{\star}}}, A^{{\hat{\star}}}, P^{{\hat{\star}}}, R^{{\hat{\star}}}, \gamma^{{\hat{\star}}})$ where
$\kappa$ is the number of colors and $[\kappa]_\textnormal{odd}\coloneqq\{k \in [\kappa] \mid k \textnormal{ is odd}\}$;
$S^{{\hat{\star}}}{=}S^\times{\times}[\kappa]_\textnormal{odd}$ is the set of multilevel product states;  $S_\mu^{{\hat{\star}}}{=}S_\mu^\times{\times}[\kappa]_\textnormal{odd}$ and $S_\nu^{{\hat{\star}}}{=}S_\nu^\times{\times}[\kappa]_\textnormal{odd}$ are the controller and adversary multilevel product states respectively;
$s_0^{{\hat{\star}}}{=}\langle s_0^\times,1\rangle$ is the initial state;
$A^{{\hat{\star}}}{=}A^\times \cup \{\beta_i \mid i>1 \textnormal{ and } i \in [\kappa]_\textnormal{odd}\}$ is the set of actions;
$P^{{\hat{\star}}}{:}S^{\hat{\star}}{\times} A^{\hat{\star}}{\times} S^{\hat{\star}}{\mapsto} \allowbreak[0,1]$ is the probabilistic transition function such that $P^{\hat{\star}}(\langle s^\times, \bar{\kappa} \rangle, a^{\hat{\star}}, \langle {s^\times}', \bar{\kappa}' \rangle ) := $
\begin{align*}
    \begin{cases}
    P^\times (s^\times, a^\times, {s^\times}') & \textnormal{if } \bar{\kappa}'{=}\bar{\kappa} \textnormal{ and } a^{\hat{\star}} {\in} A^\times(s^\times), \\
    1 & \textnormal{if } \bar{\kappa}'{>}\bar{\kappa},  a^{\hat{\star}}{=}\beta_{\bar{\kappa}} \textnormal{ and } {s^\times}'{=}s^\times{\in} S_\mu^\times, \\
    0 & \textnormal{otherwise};
    \end{cases}
\end{align*}
and $R^{\hat{\star}}:S^{\hat{\star}}\mapsto[0,1]$ and $\gamma^{\hat{\star}}:S^{\hat{\star}}\mapsto[0,1]$ are the reward and discount functions respectively such that
\begin{align}
    R_\varepsilon^{\hat{\star}}(\langle s^\times,\bar{\kappa}\rangle) &\coloneqq \begin{cases}
       \varepsilon^2 & \textnormal{if } C^\times(s^\times) = \bar{\kappa}, \\
       0 & \textnormal{otherwise},
    \end{cases} \label{eq:TMPG_rewards} \\
    \gamma_\varepsilon^{\hat{\star}}(\langle s^\times,\bar{\kappa} \rangle) &\coloneqq \begin{cases}
       1-\varepsilon & \textnormal{if } C^\times(s^\times) > \bar{\kappa}, \\
       1-\varepsilon^2 & \textnormal{if } C^\times(s^\times) = \bar{\kappa}, \\
       1-\varepsilon^3 & \textnormal{if } C^\times(s^\times) < \bar{\kappa}.
    \end{cases} \label{eq:TMPG_discount_factors}
\end{align}
\end{definition}

The TMPG $\mathcal{G}^{\hat{\star}}$ of the product game contains a copy of the product game $\mathcal{G}^\times$ for each odd color. The game starts in the first level, and in the $\bar{\kappa}$-th level, the controller can nondeterministically take a $\beta_{\bar{\kappa}'}$ action to transition to the $\bar{\kappa}'$-th level where $\bar{\kappa}'$ is an odd number larger than $\bar{\kappa}$.
In the $\bar{\kappa}$-th level, there are three distinct discount factors $1-\varepsilon$, $1-\varepsilon^2$ and $1-\varepsilon^3$ for the colors larger than $\bar{\kappa}$, the color $\bar{\kappa}$, and the colors smaller than $\bar{\kappa}$ respectively. The controller's nondeterministic decision to focus on a particular odd color is represented by taking a $\beta$-action. The restriction in a TMPG is that after focusing on an odd color, the controller can only make a decision to focus on a larger color. This prevents cyclic transitions between the levels of the TMPG.

We now show that the values of the states in a TMPG provide a lower bound for the satisfaction probabilities.

\begin{theorem}\label{theorem:TMPG_values}
For the TMPG $\mathcal{G}^{\hat{\star}}$ of a product game $\mathcal{G}^\times$, let $(\mu^{\hat{\star}},\nu^{\hat{\star}})$ be an arbitrary pure memoryless strategy pair in $\mathcal{G}^{\hat{\star}}$ and let $(\mu^\times,\nu^\times)$ be its induced strategy pair in $\mathcal{G}^\times$. Then, %
\begin{align}
    \lim_{\varepsilon\to 0^+}v^{\hat{\star}}_{\varepsilon,\mu^{\hat{\star}},\nu^{\hat{\star}}}(\langle s^\times, \bar{\kappa}\rangle) \leq Pr_{\mu^\times,\nu^\times}(s^\times \models \varphi^\times)
\end{align}
for any $s^\times\in S^\times$ and $\bar{\kappa}\in [\kappa]_\textnormal{odd}$.
\end{theorem}

\begin{proof}
Any path under any strategy pair in a TMPG~eventually reaches a level where it stays forever because there is no transition from an upper level to a lower level. %
As $\varepsilon$ goes to $0$, the return of a path approaches $1$ only if the states with the color $\bar{\kappa}$ in the $\bar{\kappa}$-th level are visited infinitely many times, and approaches $0$ otherwise. 
This can be easily shown by following similar steps as in the proof of Theorem \ref{theorem:multiple_discounts_and_rewards}. Since visiting a state colored with an odd number infinitely often while visiting the states with larger colors only finitely many times satisfies the parity condition, the values of the states provide a lower bound of
the satisfaction probabilities as $\varepsilon$ goes to $0$.
\end{proof}

Theorem~\ref{theorem:TMPG_values} implies that a controller strategy maximizing the minimum values in a TMPG achieves a lower bound on the satisfaction probabilities against an optimal adversary. This lower bound can be informally described as the probability of reaching a component in the game where the parity condition is satisfied, and the adversary cannot change the largest color among the colors of the states visited infinitely many times. In many SGs, such components exist and can be almost surely reached, making the lower bound $1$, and thus the optimal controller strategy for the TMPG induces a controller strategy in the original SG that is optimal for the given LTL~specification {as we show in a case study in Section~\ref{sec:case}.}

\section{Case Studies} \label{sec:case}
In this section, we illustrate the applicability of our approach in several case studies focused on robot navigation and control of a robotic arm. \update{We consider different LTL tasks, larger grids, continuous state-spaces, continuous action spaces, different levels of stochasticity and adversarial capabilities to evaluate the effectiveness and scalability of the methods.}
In particular, we compared the performance of our methods with the method from Hahn et al. (2020)~\cite{hahn2020}, which we refer to as the \textbf{APG} (augmented product game) method; specifically, we evaluated our reduction (referred to as \textbf{PG} -- product game), lazy color generation (referred to as \textbf{MPG} -- multilevel product game), and three-color approximation (referred to as \textbf{TMPG} -- three-color multilevel product game) methods. We did not compare with our preliminary methods from~\cite{bozkurt2021a} as those do not support general LTL formulas.

\subsection{Robot Navigation}
We start with three path-planning case studies where a robot needs to learn to navigate to perform given tasks in planar environments.

\subsubsection{Active Adversaries}
\label{sec:robot_navigation_1}
In the first case study, the environments are modeled as grids.
In these grids, each cell represents a state, and a robot can move from one cell to a neighbor cell by taking four actions: \textit{up}, \textit{down}, \textit{right} and \textit{left}. An adversary can observe the actions the robot takes and can act to disturb the movement so that the robot may move in a perpendicular direction of the intended direction. 

Specifically, there are four actions the adversary may choose \textit{none}, \textit{cw}, \textit{ccw}, \textit{both}; %
for \textit{none}, the robot will move in the intended direction w.p. $1$, %
for \textit{cw} (or \textit{ccw}), the robot will move in the intended direction w.p. $0.8$, and move in the perpendicular direction that is  \ang{90} clockwise (or counter-clockwise respectively) w.p. $0.2$; %
for \textit{both}, the robot will move in any of the perpendicular directions with w.p. $0.1$. The robot cannot leave a trap cell and if the robot attempts to move towards an obstacle or a grid edge, the robot does not move and stays in the previous cell. %

\begin{figure*}
    \captionsetup[subfigure]{aboveskip=-5pt,belowskip=5pt}
    \centering
    \begin{subfigure}[b]{0.32\textwidth}
        \centering
        \includegraphics[width=\textwidth]{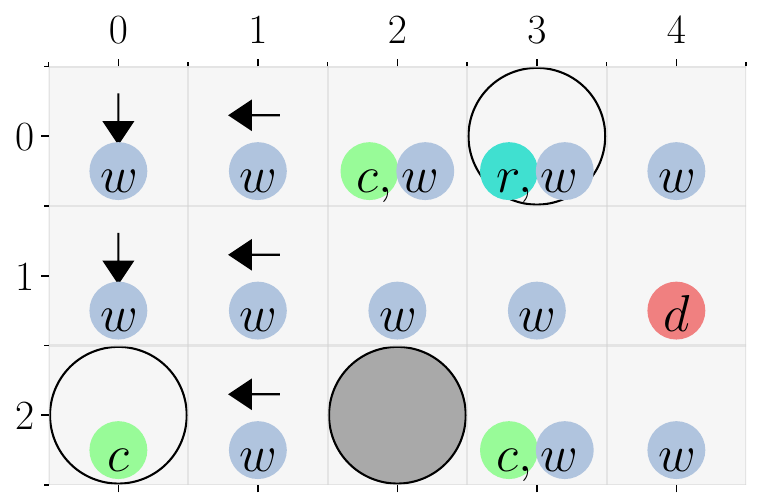}
    \end{subfigure}
    \hfill
    \begin{subfigure}[b]{0.32\textwidth}
        \centering
        \includegraphics[width=\textwidth]{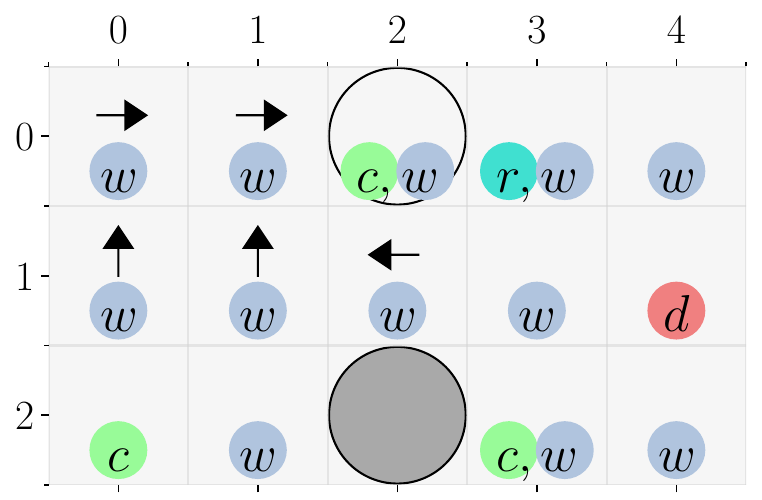}
    \end{subfigure}
    \hfill
    \begin{subfigure}[b]{0.32\textwidth}
        \centering
        \includegraphics[width=\textwidth]{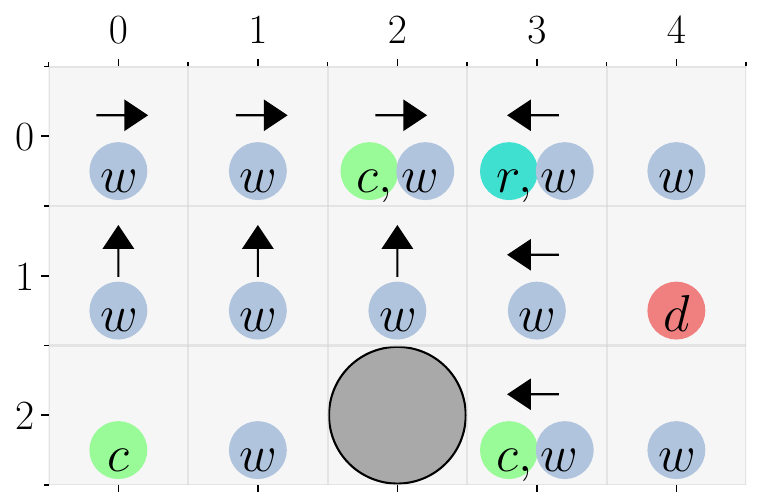}
    \end{subfigure}
    \begin{subfigure}[b]{0.32\textwidth}
        \centering
        \includegraphics[width=\textwidth]{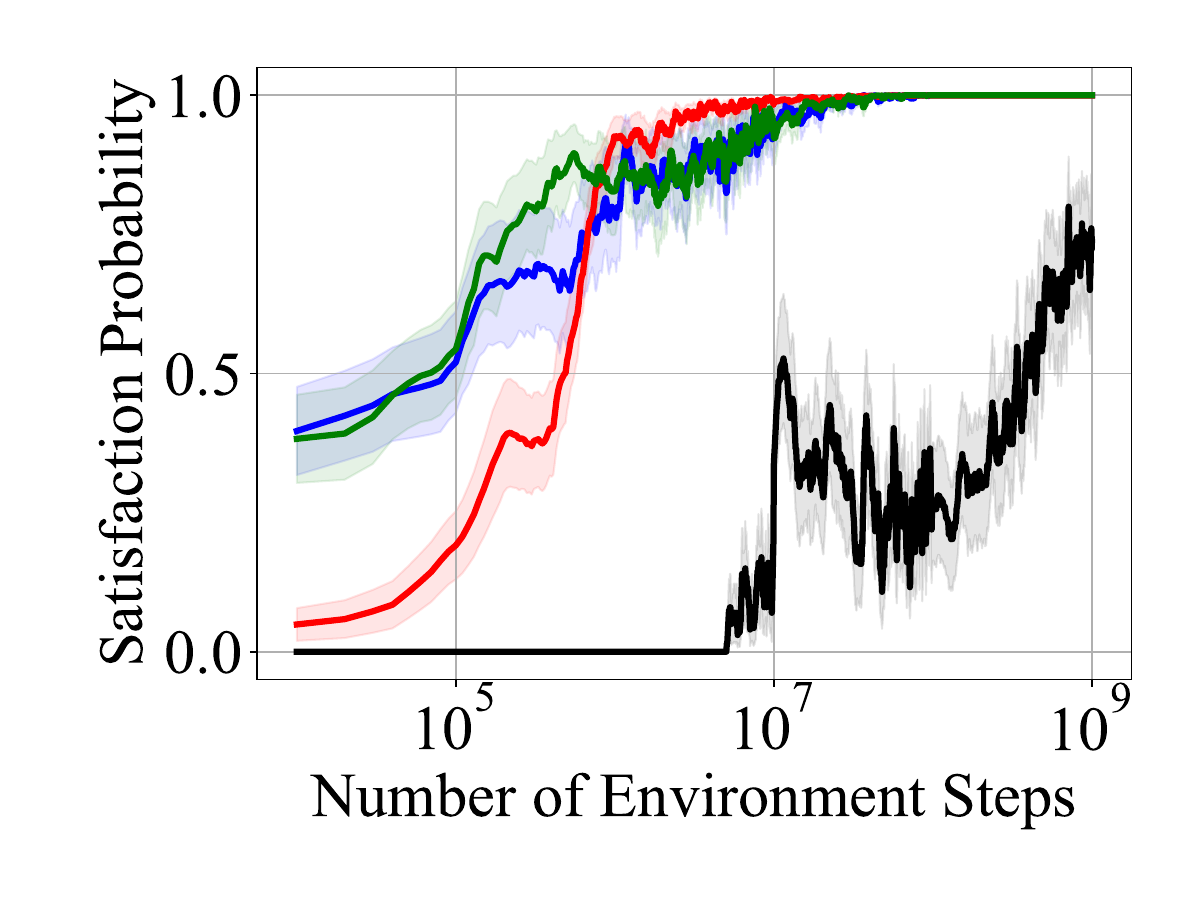}
        \caption{Environment I (Discrete)}
        \label{fig:c}
    \end{subfigure}
    \hfill
    \begin{subfigure}[b]{0.32\textwidth}
        \centering
        \includegraphics[width=\textwidth]{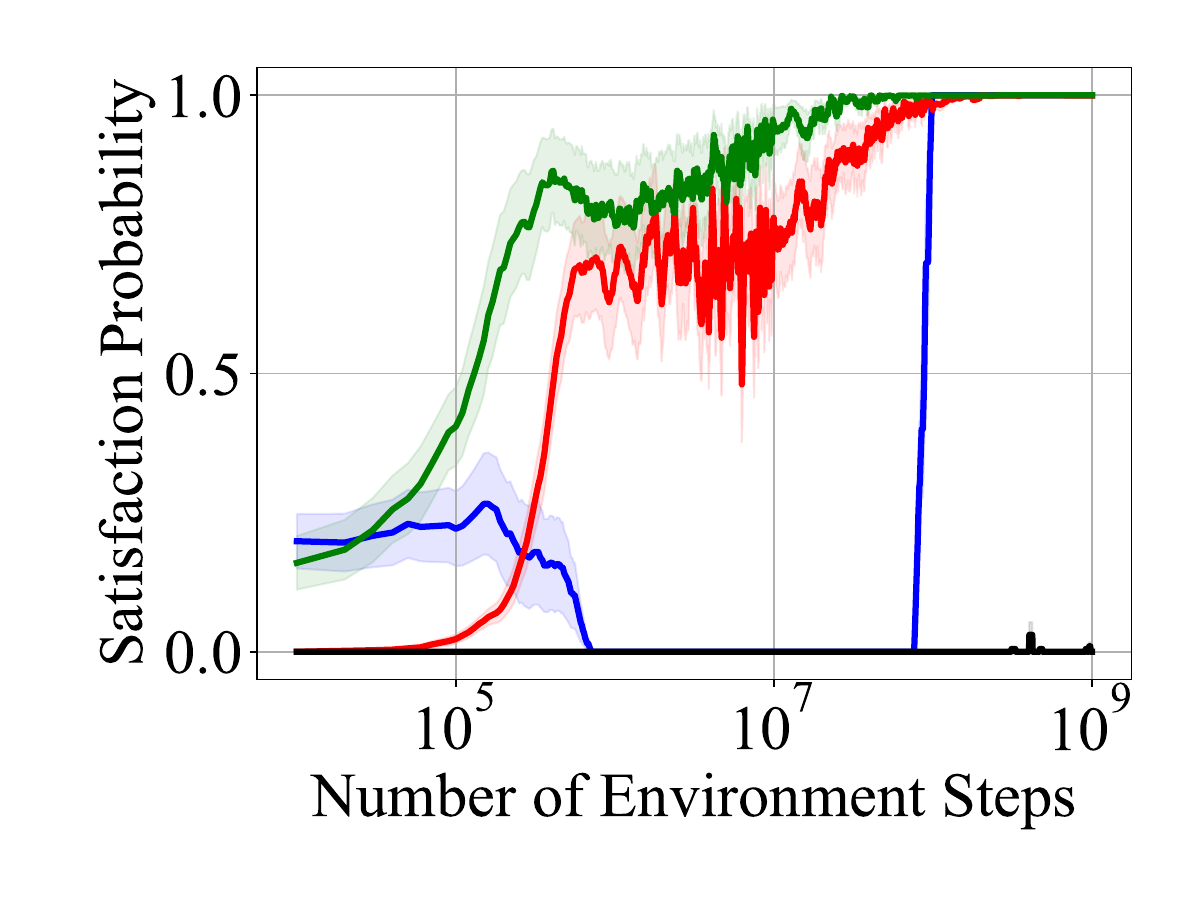}
        \caption{Environment II (Discrete)}
        \label{fig:cw}
    \end{subfigure}
    \hfill
    \begin{subfigure}[b]{0.32\textwidth}
        \centering
        \includegraphics[width=\textwidth]{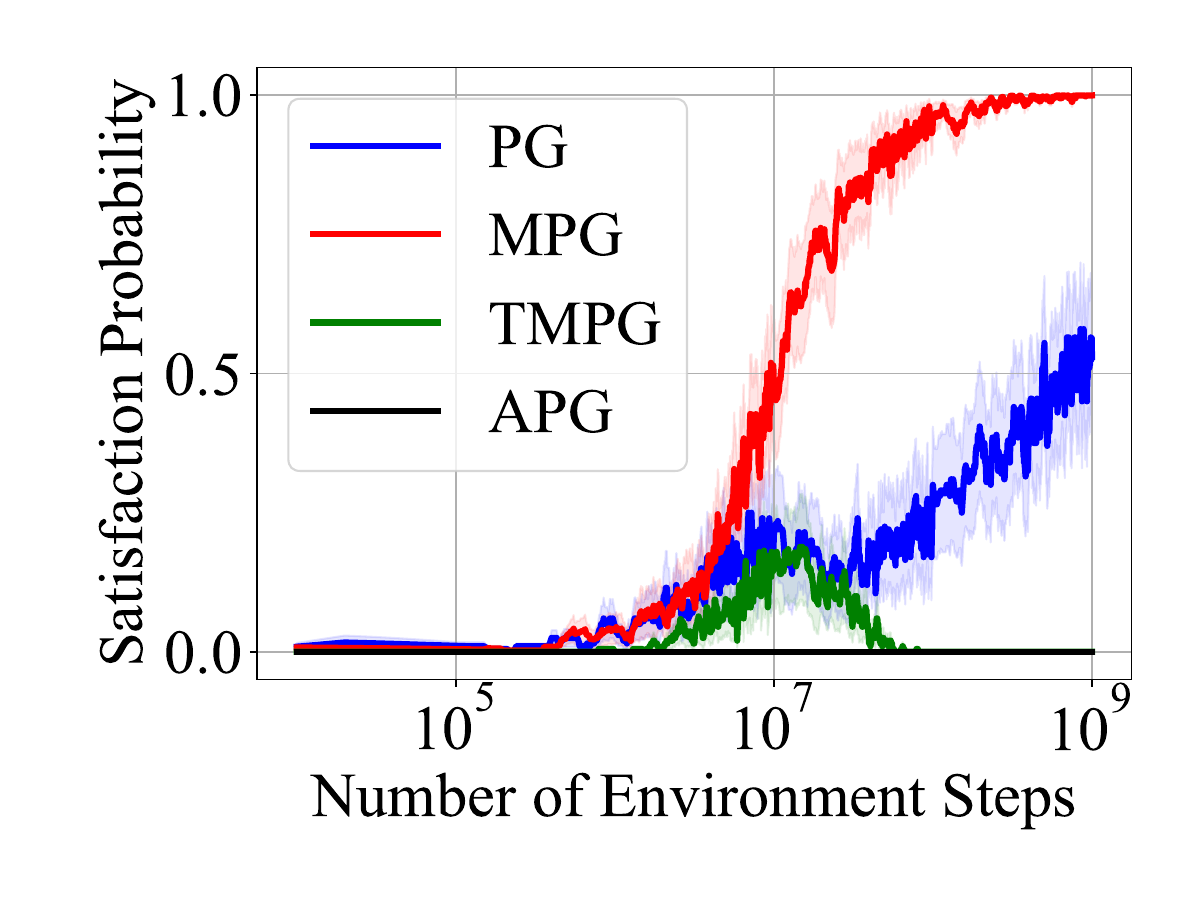}
        \caption{Environment III (Discrete)}
        \label{fig:cwrw}
    \end{subfigure}
    \begin{subfigure}[b]{0.32\textwidth}
        \centering
        \includegraphics[width=\textwidth]{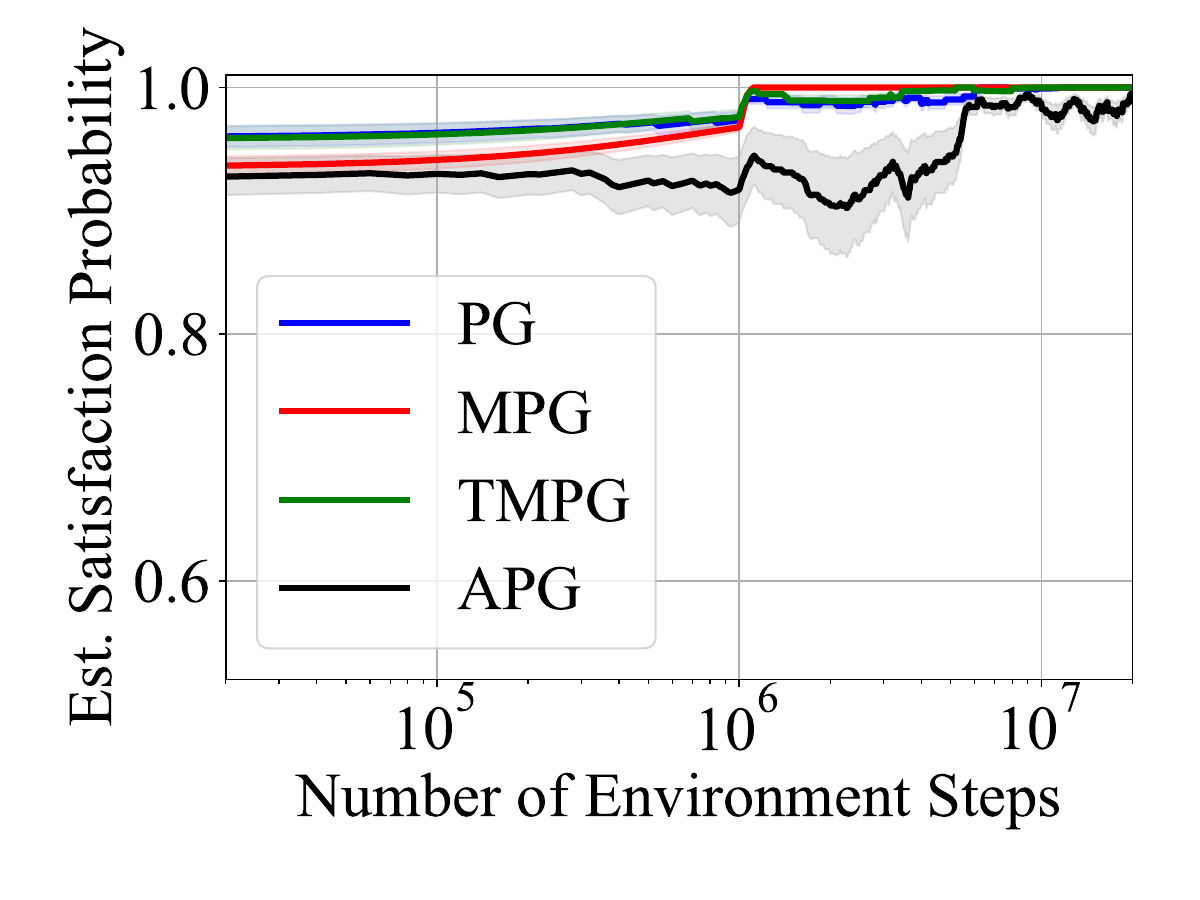}
        \caption{Environment I (Continuous)}
        \label{fig:c_cont}
    \end{subfigure}
    \hfill
    \begin{subfigure}[b]{0.32\textwidth}
        \centering
        \includegraphics[width=\textwidth]{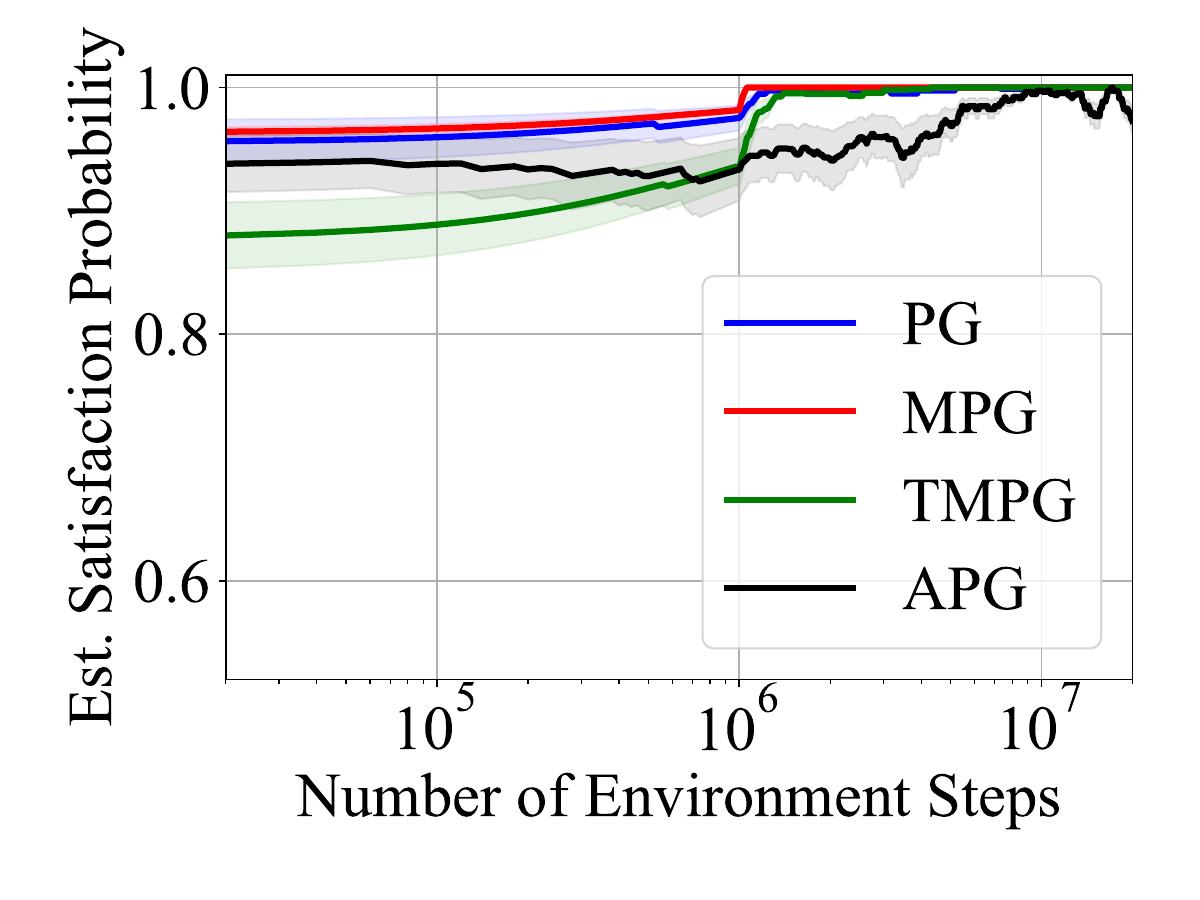}
        \caption{Environment II (Continuous)}
        \label{fig:cw_cont}
    \end{subfigure}
    \hfill
    \begin{subfigure}[b]{0.32\textwidth}
        \centering
        \includegraphics[width=\textwidth]{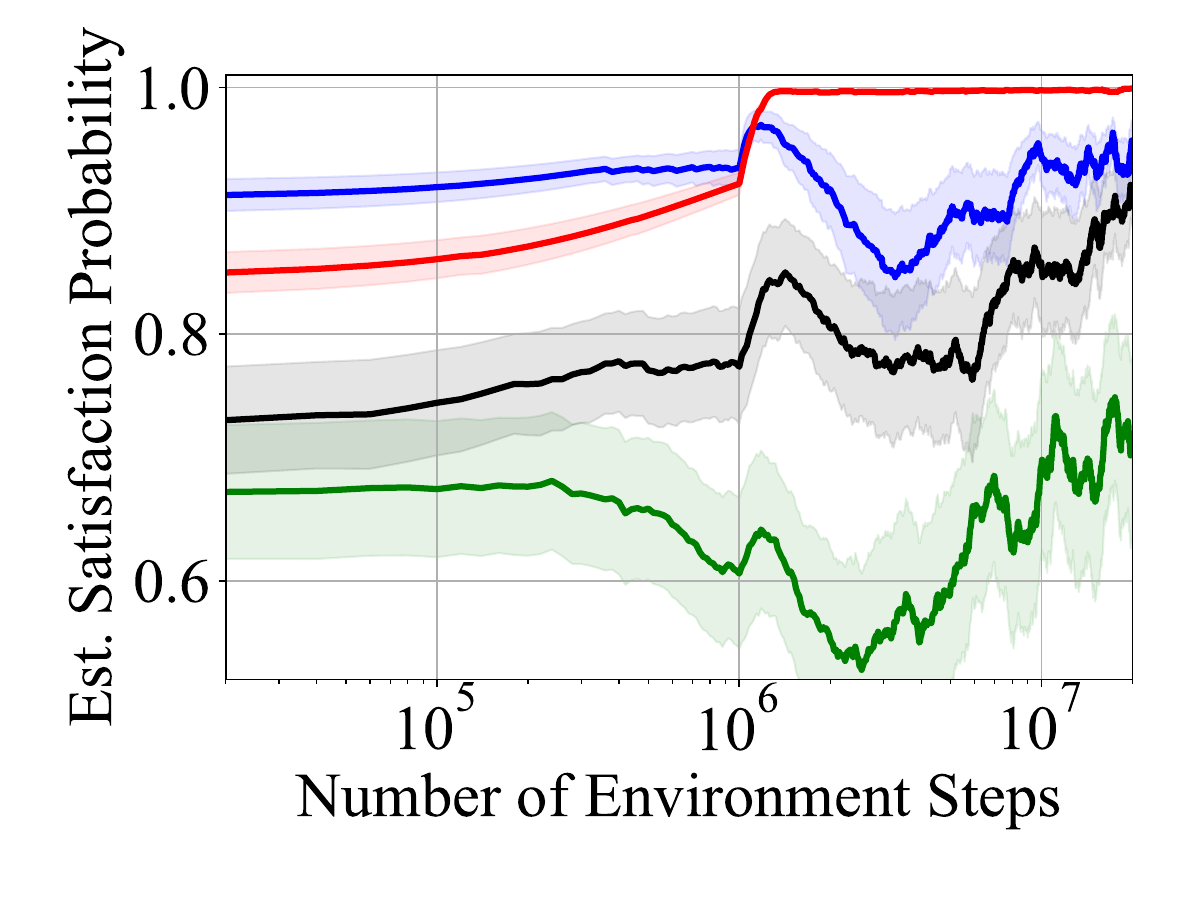}
        \caption{Environment III (Continuous)}
        \label{fig:cwrw_cont}
    \end{subfigure}
    \vspace{-6pt}
    \caption{The environments and the obtained learning curves for the case studies in Sections \ref{sec:robot_navigation_1} and \ref{sec:robot_navigation_2}. The shaded regions are the quarter of the standard deviations, and the results are smoothed by moving averages for better visualization. In the environments, the encircled letters are labels, the empty circles are trap cells; the filled circles are obstacles, and the arrows are optimal controller actions. The blue, red, green, and black learning curves for the PG, MPG, TMPG, and APG methods.}%
    \label{fig:robot_navigation}
\end{figure*}

Fig.~\ref{fig:robot_navigation} captures the considered scenarios. 
The trap cells are represented as large empty circles, and the obstacles are represented as large circles filled with gray. The labels of the cells (i.e., states) are displayed as encircled letters filled with various colors.

The core navigation task of the robot is to enter the workspace ($w$) and stay there while repeatedly charging ($c$) and monitoring the assigned region ($r$). Alternatively, the robot can go to and stay in a charging station ($c$). In addition, the robot must always avoid the danger zone ($d$). This entire task can be formally captured as the LTL~formula: 
\begin{align}
    \varphi_1 \coloneqq \big(\left(\lozenge \square w \wedge \square \lozenge c \wedge \square \lozenge r  \right) \vee \lozenge \square c \big) \wedge \square \neg d. \label{eq:varphi_1}
\end{align}

Fig.~\ref{fig:robot_navigation} illustrates the three ways of performing the task considered in this case study:
\begin{itemize}
    \item (I) reaching and staying in the charging station at $(2,0)$, which is not in the workspace;
    \item (II) reaching and staying in the charging station at $(0,2)$ in the workspace;
    \item (III) visiting repeatedly the charging station at $(0,2)$ and the assigned region at $(0,3)$ without leaving the workspace.
\end{itemize}
We used \emph{Owl}~\cite{owl} to automatically translate LTL specifications into DPAs.
The DPA translated from $\varphi_1$ has $4$ states, and $5$ colors (\ifx\arxiv\undefined Fig.~\ref{fig:dpas}\else Fig.~\ref{fig:varphi_1}\fi). %
The largest colors among the colors of the DPA transitions made infinitely often while performing (I), (II), and (III) are $3$, $1$, and $3$ respectively.

We used minimax-Q~\cite{littman1994} to learn the optimal controller strategies. %
We set $\varepsilon$ to $0.01$ and $\tau_{\varepsilon}$ to $\sqrt{\varepsilon}=0.1$.
During the learning phase, an $e$-greedy strategy was followed, which chooses random actions w.p. $e$ and chooses greedily otherwise. We initially set $e=1$ and gradually decrease it to $0.1$ to encourage exploration as both the controller and the adversary need to actively search for optimal strategies against each other. We gradually decreased the learning rate $\alpha$ with the number of steps from $0.5$ to $0.05$. Each episode started in the cell at $(0,0)$ and terminated after a $\mathcal{T}=10^3$ time steps.
We evaluated the performance of the learned controller strategies against the optimal adversary strategies. We formally derived the optimal adversary strategies and the associated minimum satisfaction probabilities from the full specification of the underlying SG  using the \emph{PRISM} model checking tool~\cite{prism}. The learning curves of the methods for each environment are shown in the second row of Fig.~\ref{fig:robot_navigation} where the probabilities are averaged over $10$ simulations.

In the first environment, learning to perform (I) is a simple task as the cell $(2,0)$ containing the charging station aimed to be reached is a trap cell, and the corresponding color $3$ is large (yielding smaller discount factors) relative to the largest color $4$. Thus, MPG and TMPG do not have any advantage over PG. However, the performance of MPG and TMPG is not significantly less than the one of PG despite the increased space complexity as shown in Fig.~\ref{fig:c}. APG performs poorly compared to these methods as the rewards provided by APG are very sparse.

In the second environment, learning to perform (II) is harder since the corresponding color is $1$, resulting in large discount factors. TMPG outperforms the other methods as (II) can be performed by merely focusing on the color $3$. The performance of MPG is significantly better than the ones of PG and APG as the task can be performed in the second level. However, the performance is slightly less than TMPG because of the transitions to the upper levels that can happen with a small probability. PG could not converge until $10^8$ steps due to the large discount factors. APG did not converge even with $10^9$ steps as the probability of getting a positive reward is $10^{-8}$.

In the third environment, learning to perform (III) is considerably harder as two cells need to be repeatedly visited without reaching the danger zone in the presence of adversarial actions. The controller must take the action \textit{left} in the cells $(0,3)$, $(1,3)$, and $(2,3)$ because otherwise, the adversary can eventually pull the robot to the danger zone. MPG significantly outperforms the other methods as shown in Fig.~\ref{fig:cwrw}, and PG is slowly converging. TMPG failed to converge as this task requires two colors, $3$ and $1$ to be focused, corresponding to (II), and the reaching and staying in the cell $(2,3)$. If the controller focuses on $3$, the adversary can pull the robot to the cell $(2,3)$, which requires focusing on $1$, and if the controller focuses on $1$, the adversary can pull the robot out of $(2,3)$. APG, in this case as well, failed to converge.

Overall, our MPG method outperforms our methods PG and TMPG, as well as the existing method APG for a general LTL task. TMPG performs better than MPG if the task can be performed by focusing on only one color; however, it could perform poorly otherwise. Finally, the existing APG method is significantly outperformed by our methods due to its prohibitively sparse rewards.

\update{
\subsubsection{Continuous Environments}
\label{sec:robot_navigation_2}
In this case study, the environments are the continuous version of the environments in the first case study. The state space is defined as the continuous (y,x)-position of the robot where the origin is at the top left corner and the positive directions of y and x are downwards and rightwards respectively. Similar to the first case study, the robot can take four actions to move in four directions. The robot moves $1$ unit in the intended direction perturbed by bivariate Gaussian noise with a variance of $\sigma^2 I$ where $\sigma$ is $0.05$ and $I$ is a $2\times2$ identity matrix. The noise is truncated to fit in a circle with a radius of $4\sigma$.

The adversary is located within the obstacle at $(2.5, 2.5)$, and can take the actions \textit{push} and \textit{pull} to move the robot towards or away from the obstacle. Specifically, let $(y,x)$ denote the position of the robot; then the adversary can perturb the x-position of the robot if $y$ is in $[2,3]$ and can perturb the y-position of the robot if $x$ is in $[2,3]$. The amount of perturbation is drawn from a uniform distribution over the interval $[0,2]$. Similar to the previous case study, the robot cannot leave a trap region, cannot go inside the obstacle, and is constrained by the borders of the environment; if the robot attempts, the robot moves and remains on the borders or the edges of the obstacle.
A position $(y,x)$ in the continuous versions of the environments has the characteristics of the corresponding grid cell; e.g., same label, being a trap or an obstacle. The robot needs to learn how to map these continuous positions to actions to perform the LTL task $\varphi_1$ from \eqref{eq:varphi_1} against any adversary~strategy. 

We adopted deep RL techniques to learn the controller and adversary strategies. We modified Deep Q-Network (DQN) \cite{mnih2015,sb3} to include adversarial actions and minimax-Q updates. We used the default parameters with two hidden layers of size $64$. We set $\varepsilon=0.25$ to increase discounting and thereby stability as performing $\varphi_1$ in these environments does not require long-horizon planning. 

In this case study, the optimal adversary strategies and the corresponding minimum satisfaction probabilities cannot be derived using the model checking tools such as \emph{PRISM} \cite{prism} due to the continuous state-space and the nonlinear dynamics. Instead, in our evaluation, we used a manually crafted optimal adversary strategy 
that is to push the robot if the robot is on the right-hand side of the obstacle, and to pull otherwise. 

In addition, since we cannot directly compute the minimum satisfaction probabilities for the learned controller strategies (as that would require reasoning on infinite traces), we estimated them as follows. We considered an episode satisfying if the largest color in the second half of the episode is an odd number and occurs at least $20\%$ of the time (a proxy for satisfaction of repeated reachability condition) -- for the environments shown in Fig.~\ref{fig:robot_navigation}, it is reasonable to assume that, on average, one of five transitions will be colored with the largest odd number visited. The results, however, are not sensitive to this ratio. 
We calculated the estimated satisfaction probability as the ratio of the number of satisfying paths to the number of all evaluation episodes, and used $1000$ evaluation episodes of length~$\mathcal{T}=100$.

The learning curves of the methods for each environment against the optimal adversary are shown in the third row of Fig.~\ref{fig:robot_navigation}; the estimated satisfaction probabilities are averaged over $4$ simulations. All of our methods converged to an optimal controller strategy around $10^6$ steps in the first and the second environments while APG struggled to converge about until $10^7$ steps and exhibited high variance. In the last environment, MPG outperformed the other methods. PG was able to quickly learn a near-optimal controller strategy; however, it failed to converge to an optimal one. Similar to the first case study, TMPG performed worse than the rest in this environment.
 
\subsubsection{Long-Horizon Planning}
\label{sec:robot_navigation_3}
This case study considers a larger grid to evaluate the performance of the methods for long-horizon planning. Here, the robot can deterministically move to an adjacent cell using the same four actions, and the adversary can manipulate the position of the garbage area ($g$) and the danger zone ($d$) within the designated regions. Fig.~\ref{fig:large_grid} shows the environment and the designated regions considered in this case study. The initial position of the garbage area is $(2,6)$, and the adversary can change it to one of $(2,5)$, $(2,6)$ and $(2,7)$ at any time step w.p. $0.1$. Similarly, the initial position of the danger zone is $(5,1)$, and the adversary can change it to one of $(5,0)$, $(5,1)$ and $(5,2)$ at any step~w.p.~$0.1$.

\begin{figure}[!t]
    \centering
    \includegraphics[width=\columnwidth]{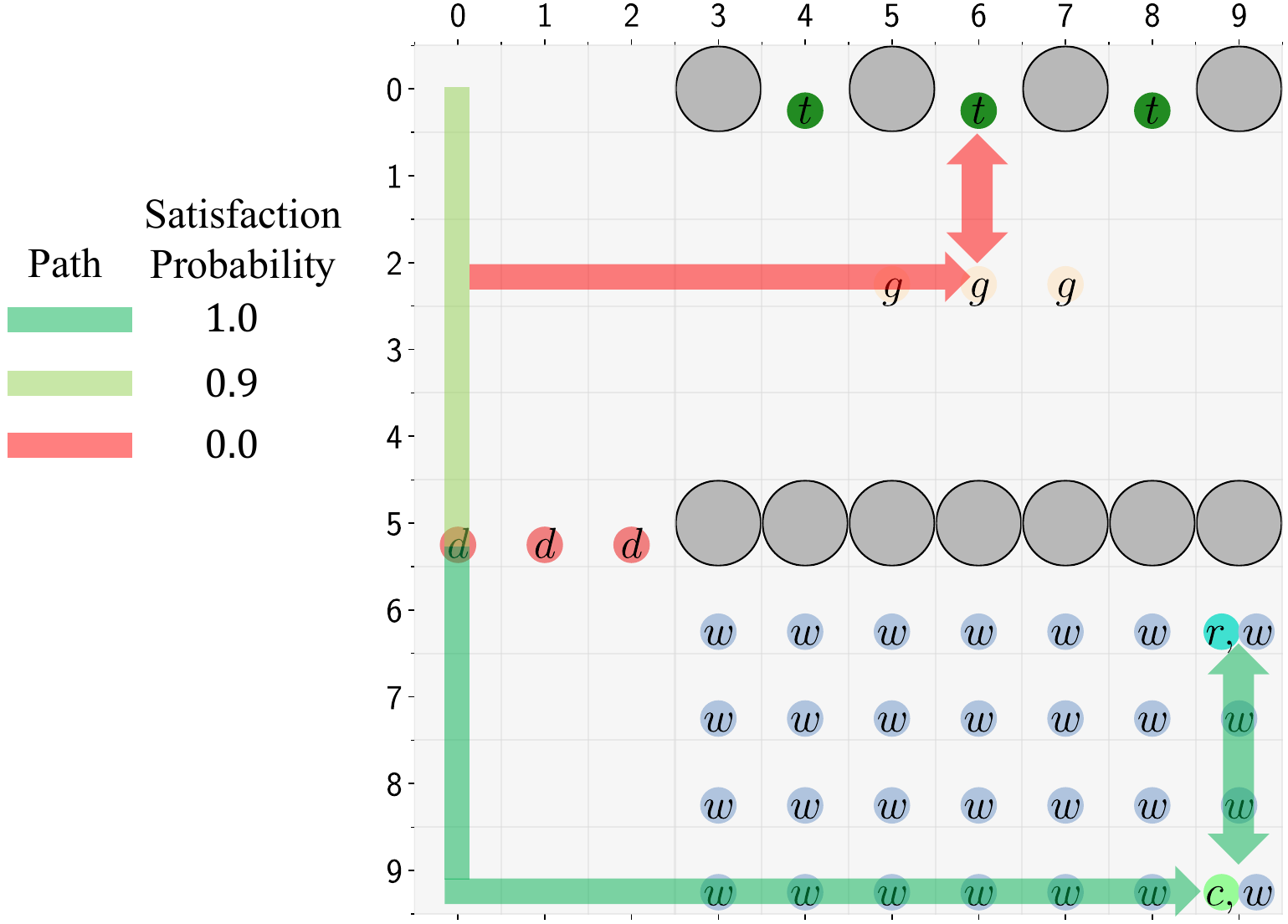}
    \caption{The environment for the case study in Section \ref{sec:robot_navigation_3}. The encircled letters are labels, the empty circles are trap cells; the filled circles are obstacles. The arrows represent two ways of satisfying the LTL task $\varphi_2$ from \eqref{eq:varphi_2} where the colors of the arrows represent the satisfaction probabilities.}
    \label{fig:large_grid}
\end{figure}

The robot needs to either pick up the garbage ($g$) and trash it out ($t$) in at most two time steps, or repeatedly monitor the assigned region ($r$) and go to the charging station ($c$). The robot can be formally defined as the following LTL formula:
\begin{align}
    \varphi_2 \coloneqq \Big( &\square \big(\lozenge g \wedge [g \to (\bigcirc t \vee \bigcirc \bigcirc t)]\big) \notag \\
    &\vee \big(\lozenge \square w \wedge \square \lozenge c \wedge \square \lozenge r  \big) \Big) \wedge \square \neg d. \label{eq:varphi_2}
\end{align}
The DPA obtained from $\varphi_2$ has $9$ states, and $4$ colors~
(Fig.~\ref{fig:varphi_2}\ifx\arxiv\undefined \ in our corresponding technical report \cite{arxiv_version}\fi).

There are two possible ways to perform this task: $(i)$~the robot can repeatedly pick up the garbage at $(2,6)$ and trash out at $(0,6)$, or $(ii)$ potentially go through the danger zone and repeatedly visit the assigned region at $(6,9)$ and charging station at $(9,9)$. Against an optimal policy, $(i)$ cannot be satisfied because the adversary will eventually move the garbage area to either $(2,5)$ or $(2,7)$ from which the robot needs to take at least $3$ time steps to trash it out. However, the robot can satisfy $(ii)$ if the robot safely passes the designated region for the danger zone. The optimal controller strategy is, therefore, to try to pass the region while avoiding the danger zone and satisfy $(ii)$; and the optimal adversary strategy is to move the danger zone to the cell the robot uses to pass the fifth row. As a result, the task can be almost surely satisfied in the lower part of the grid, and can be satisfied w.p. $0.9$ in the upper part.

We used minimax-Q to learn the optimal controller strategies using the same parameters from the first case study. We conducted three experiments where the robot starts~in~cells: $(0,0)$, $(9,0)$ and $(9,9)$ and navigates for $\mathcal{T}=1000$ environment steps. Fig.~\ref{fig:robotnavigation_long_horizon} shows the derived learning curves; the satisfaction probabilities are averaged over $10$ simulations.

MPG learned a near-optimal controller strategy for the case where the robot starts in $(0, 0)$, which is far from the assigned region and the charging station. PG was able to learn a controller strategy satisfying the task w.p. around $0.5$; however, it could not converge. The other methods failed to learn any reasonable strategy. We note that the maximum satisfaction probability against the optimal adversary is $0.9$ in this case.
For the case where the robot starts in $(9, 0)$; i.e., the robot does not need to pass through the danger zone, MPG converged to an optimal strategy faster than the other methods. PG converged slower than MPG, and TMPG learned a reasonable strategy only after $10^8$ steps. APG could not learn any strategy that yields a positive satisfaction probability.

For the last case where the robot starts in the charging station at $(9,9)$, PG learned faster than the other methods although it struggled to stably converge to an optimal strategy. MPG converged to an optimal almost as fast as PG. TMPG learned an optimal strategy only after about $10^8$ steps. APG, like in the other cases, could not learn any meaningful strategy.
Overall, MPG significantly outperformed the other methods and APG could not learn any controller strategy that satisfies the task with a positive probability.

\begin{figure*}
    \captionsetup[subfigure]{aboveskip=-15pt,belowskip=5pt}
    \centering
    \begin{subfigure}[b]{0.32\textwidth}
        \centering
        \includegraphics[width=\textwidth]{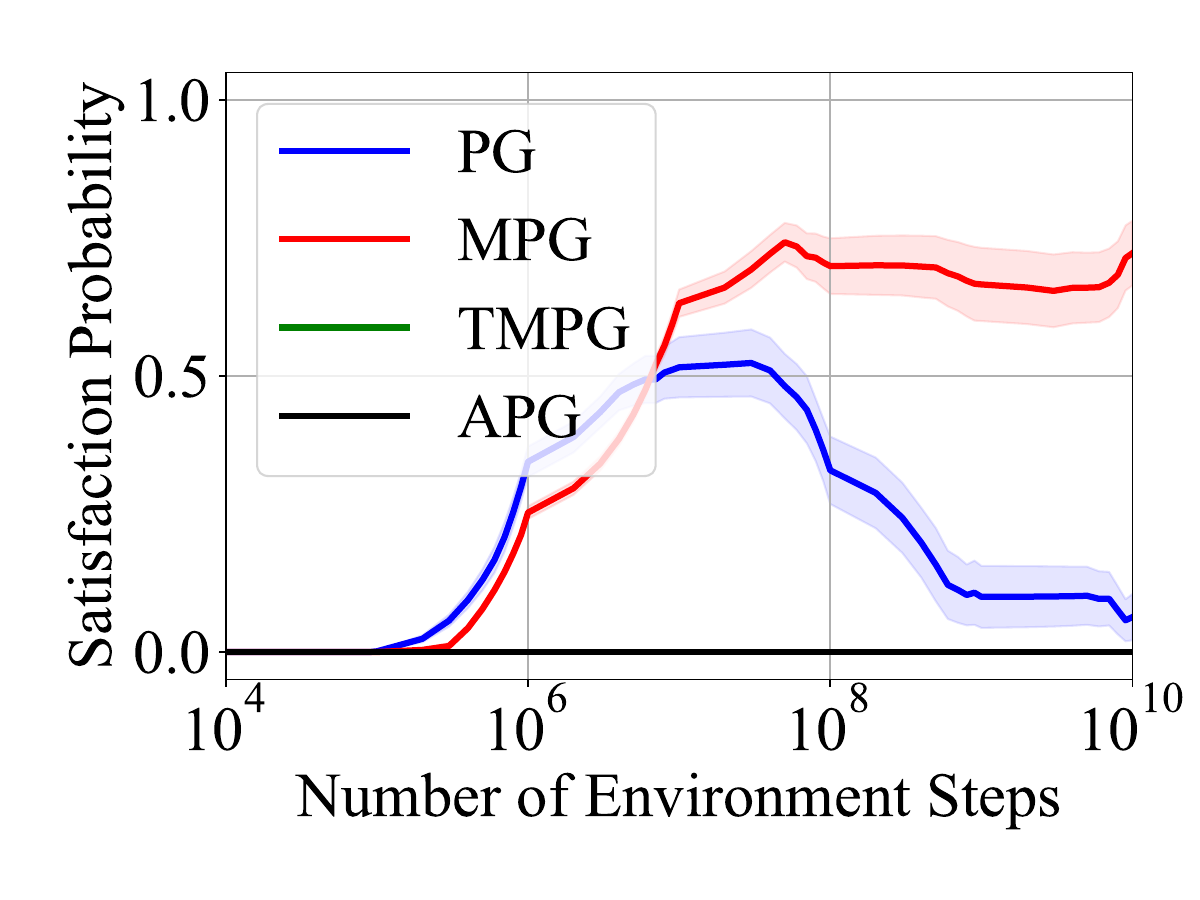}
        \label{fig:robotnavigation_(0, 0)}
        \caption{Start Cell: $(0,0)$}
    \end{subfigure}
    \hfill
    \begin{subfigure}[b]{0.32\textwidth}
        \centering
        \includegraphics[width=\textwidth]{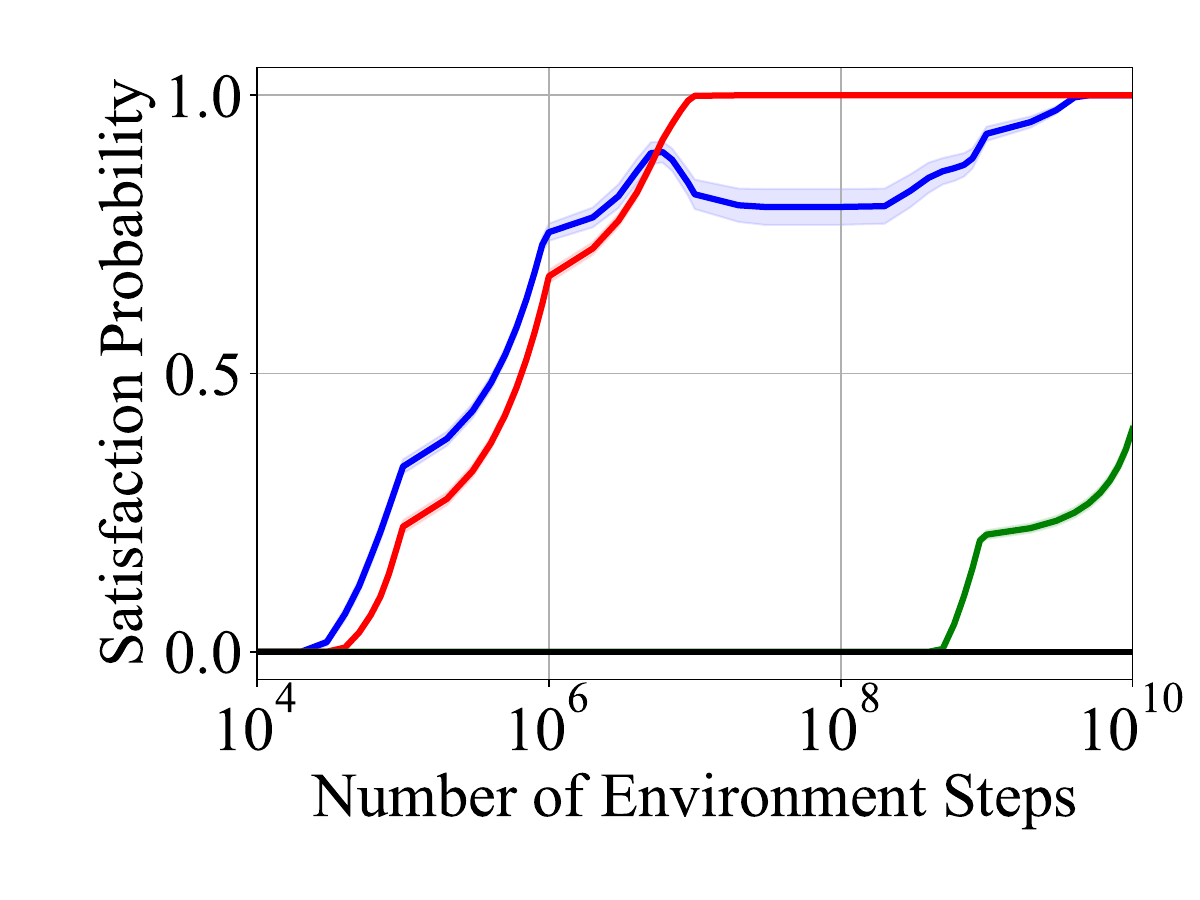}
        \label{fig:robotnavigation_(9, 0)}
        \caption{Start Cell: $(9,0)$}
    \end{subfigure}
    \hfill
    \begin{subfigure}[b]{0.32\textwidth}
        \centering
        \includegraphics[width=\textwidth]{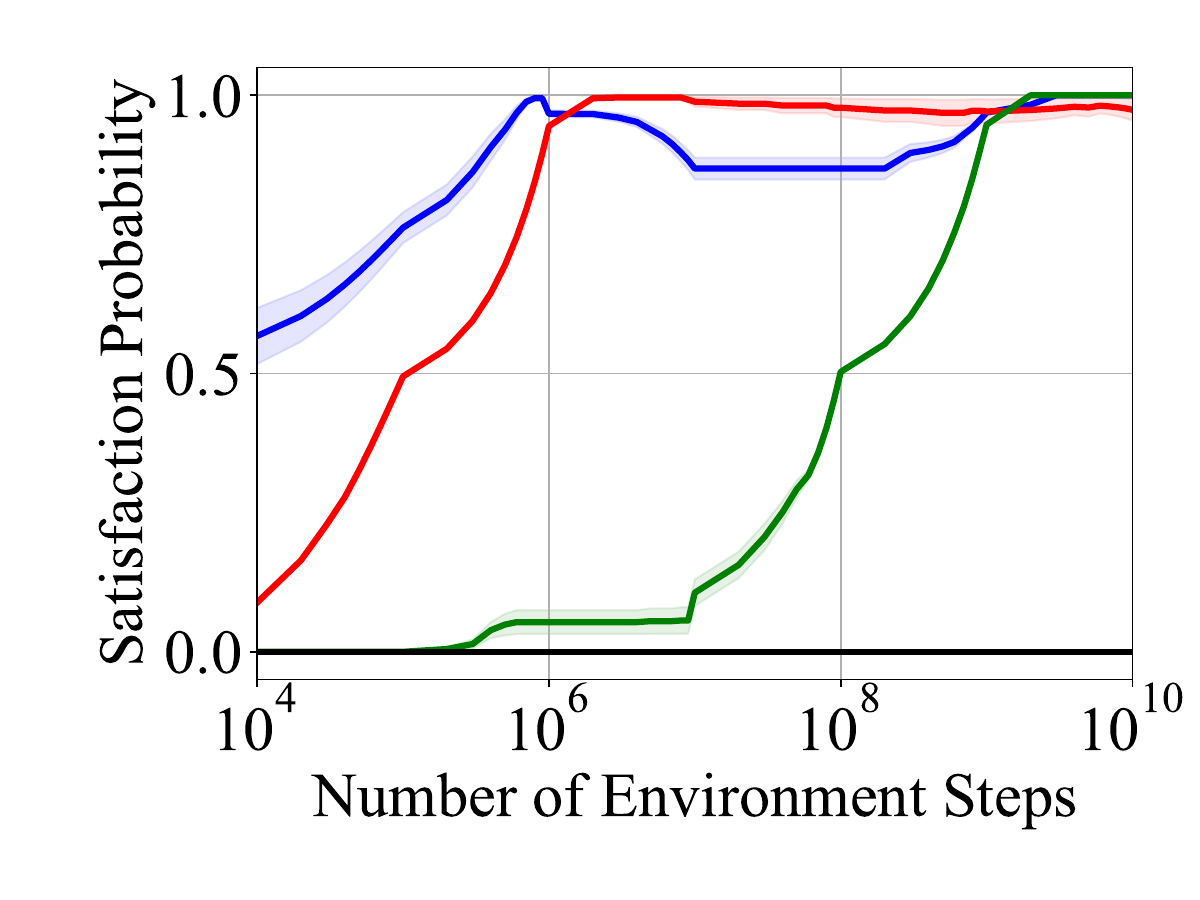}
        \label{fig:robotnavigation_(9, 9)}
        \caption{Start Cell: $(9,9)$}
    \end{subfigure}
    \vspace{-8pt}
    \caption{The learning curves for the case study in Section \ref{sec:robot_navigation_3}. The shaded regions are the quarter of the standard deviations, and the results are smoothed by moving averages for better visualization. }
    \label{fig:robotnavigation_long_horizon}
\end{figure*}
}

\subsection{Robotic Arm}
\update{
\label{sec:robotic_arm}
We consider two motion planning tasks where a robotic arm should repeatedly reach some target ball positions where the adversary can manipulate the position of one of the balls.
We implemented a simulation environment based on the Fetch environments \cite{fetch} in OpenAI Gym \cite{gym}.
In this environment, the state space consists of the positions of the gripper and the balls.
The action space of the controller is continuous and consists of three dimensions that represent the position that the arm intends to move the gripper to. In the beginning of each episode, the balls are placed in random positions that are at least $10~cm$ apart and within the reach of the arm, including the air. The red ball is placed outside the boundaries of the smallest cuboid containing the blue and green balls; however, the adversary can move the red ball to any point in the state space w.p. $0.1$.
The environment is illustrated in Fig. \ref{fig:robotic_arm_picture}.

The first task is repeatedly visiting two given random positions, represented by a green and a blue ball while staying within the boundaries of the cuboid of these balls. The robotic arm does not need to be within the cuboid if it can repeatedly reach both of the balls at the same time. The robotic arm should stay away at all costs from the red ball, which is controlled by the adversary.

The task can be represented as the following LTL formula:
\begin{align}
    \small
    \varphi_3 = \Big(\big(&\square \lozenge \textsf{green\_ball} \wedge \square \lozenge \textsf{blue\_ball} \wedge \lozenge \square \textsf{boundary} \big) \notag \\
    \vee \big(&\square \lozenge (\textsf{green\_ball} \wedge \textsf{blue\_ball}) \big) \wedge \square \neg \textsf{red\_ball}\Big), \label{eq:varphi_3}
\end{align}
which is translated to a DPA with $3$ states and $4$ colors (Fig.~\ref{fig:varphi_3}\ifx\arxiv\undefined \ in our technical report \cite{arxiv_version}\fi). The optimal adversary strategy for this task is to try to put the red ball in the position of either the green ball or the blue ball. In this way, the adversary can prevent the controller performing this task w.p. $0.1$.

The second task includes reaching and staying in the position of the red ball outside the cuboid formed by the other balls. If the red ball is in the same position as the green or the blue ball, then reaching the red ball is sufficient; the robot does not need to stay in its position or to be outside the cuboid. Alternatively, the robot can repeatedly reach a position where both of the green and the blue balls are in, but the red ball is not. This task can be formally defined as the LTL formula:
\begin{align}
    \small
    \varphi_4 = \Big(\lozenge \square &\big(\textsf{red\_ball} \wedge \neg \textsf{boundary} \big) \notag \\
    \vee \ \lozenge &\big(\textsf{red\_ball} \wedge \textsf{green\_ball}\big) \vee \lozenge\big(\textsf{red\_ball} \wedge \textsf{blue\_ball}\big) \notag \\
    \vee \ \square \lozenge &\big(\textsf{green\_ball} \wedge \textsf{blue\_ball} \wedge \neg\textsf{red\_ball}\big)\Big), \label{eq:varphi_4}
\end{align}
which is translated to a DPA with $2$ states and $4$ colors (Fig.~\ref{fig:varphi_4}\ifx\arxiv\undefined \ in our report \cite{arxiv_version}\fi). The optimal adversary strategy, in this task, is to try to move the red ball to a position that is away from the green and the blue balls but inside the boundaries of the cuboid of these balls. Against such an adversary strategy, the maximum probability that the robot successfully performs this task is~$0.9$.

\begin{figure*}
    \captionsetup[subfigure]{aboveskip=1pt,belowskip=5pt}
    \centering
    \begin{subfigure}[b]{0.3\textwidth}
        \centering
        \includegraphics[width=0.76\textwidth]{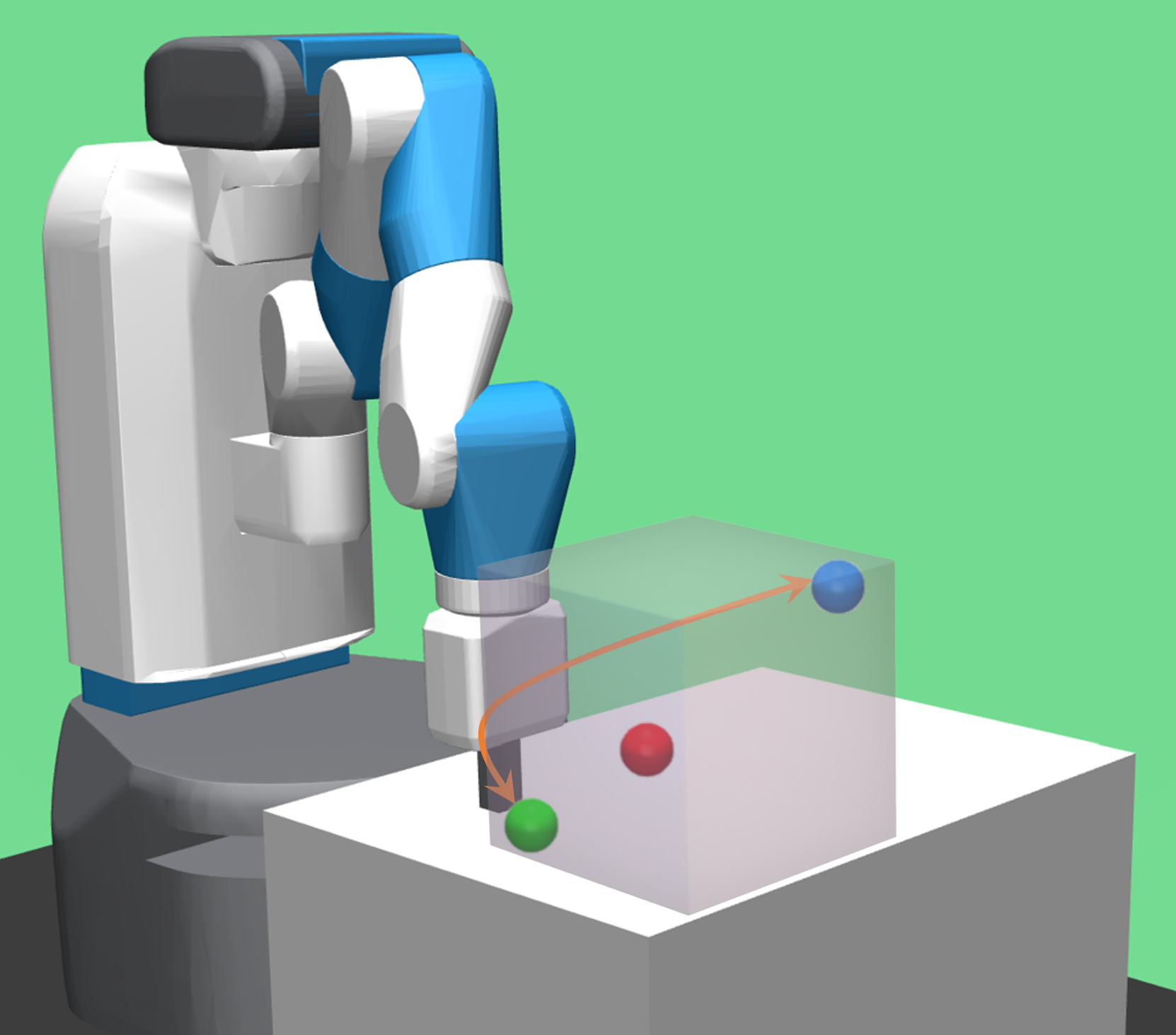}
        \caption{A robot trying to repeatedly reach the green and the blue balls while staying in the rectangular boundary and avoiding the red ball placed by the adversary.}
        \label{fig:robotic_arm_picture}
    \end{subfigure}
    \hfill
    \begin{subfigure}[b]{0.32\textwidth}
        \centering
        \includegraphics[width=\textwidth]{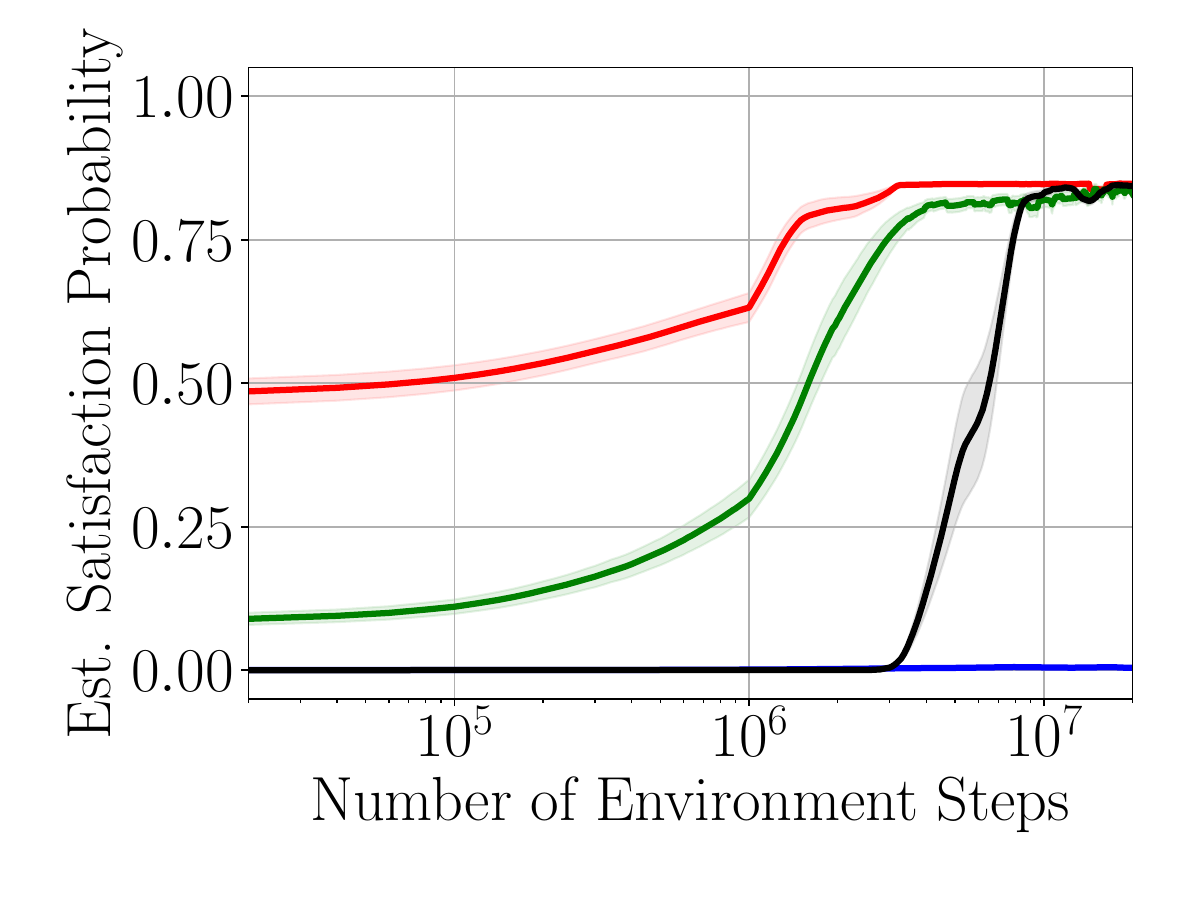}
        \caption{Task 1}
        \label{fig:roboticarm_task0}
    \end{subfigure}
    \hfill
    \begin{subfigure}[b]{0.32\textwidth}
        \centering
        \includegraphics[width=\textwidth]{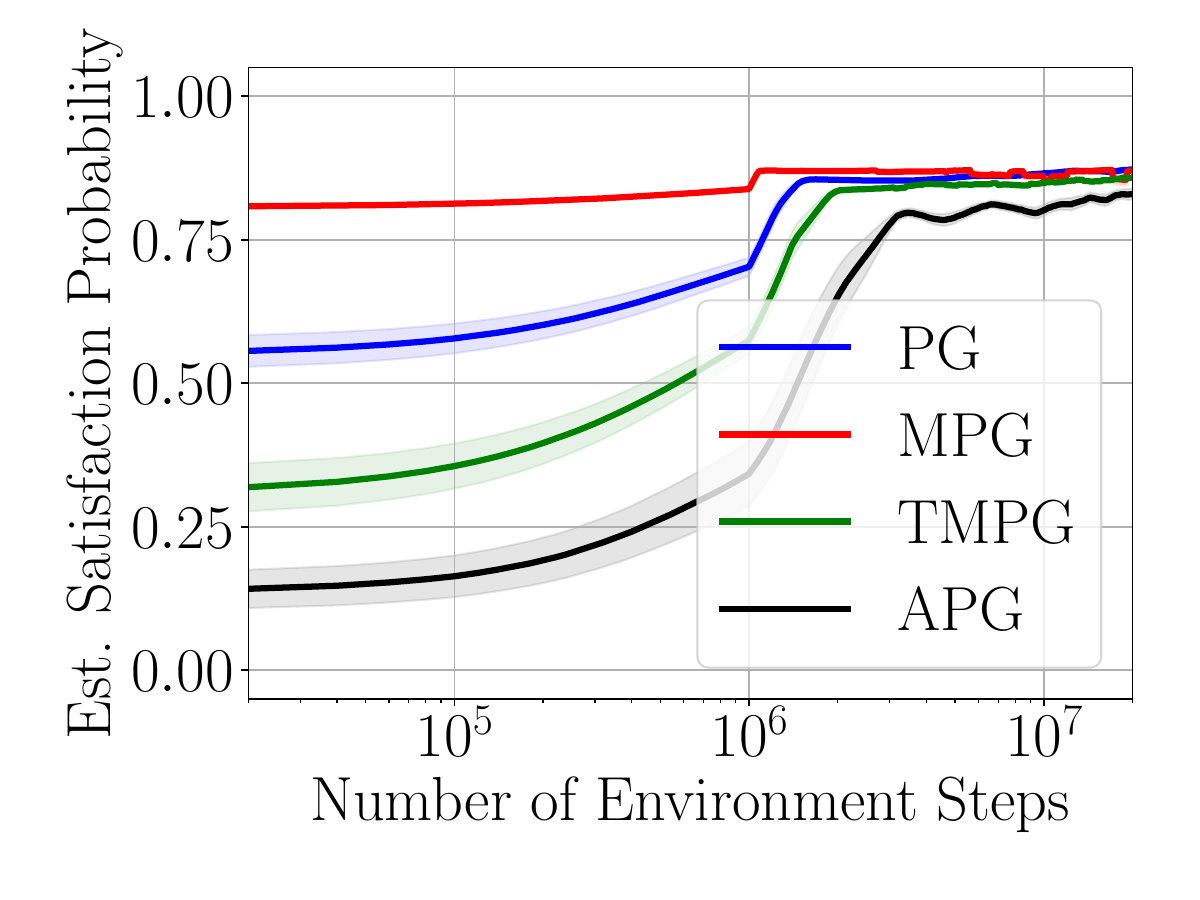}
        \caption{Task 2}
        \label{fig:roboticarm_task1}
    \end{subfigure}
    \caption{The environment and the obtained learning curves for the robotic arm tasks in Section \ref{sec:robotic_arm}. The shaded regions are the quarter of the standard deviations, and the results are smoothed by moving averages for better visualization.}
    \label{fig:roboticarm}
\end{figure*}

We integrated our framework into Truncated Quantile Critics (TQC) \cite{kuznetsov2020, sb3}, an off-policy RL algorithm for continuous actions. We used a neural network with two hidden layers of size $64$ for both the actor and the critic. Similar to the case in Section~\ref{sec:robot_navigation_2}, we used $\varepsilon=0.25$ to increase discounting and stability since the robotic arm can quickly reach any position in this environment. We used two separate TQC for the controller and the adversary. The adversary gets the cumulative discounted rewards the controller obtained throughout an episode as the reward signal.
During the learning, each episode starts in a random state, and the adversary tries to move the red ball. The controller observes the new position of the red ball and takes actions for $\mathcal{T}=100$ environment steps to learn to perform the task. We trained the controller and the adversary in turns consisting of $10^4$ exploration steps.

Similar to the case study in Section~\ref{sec:robot_navigation_2}, we manually crafted the optimal adversary strategies and estimated the minimum satisfaction probabilities via simulation as deriving them via model checking tools is not feasible. 
We evaluated the performance by following the deterministic version of the controller strategy for $1000$ evaluation episodes of length $100$ against the optimal adversary strategy. 
To estimate the satisfaction probabilities for the first task, 
we calculated the ratio of the evaluation episodes that visited both of the green and the blue balls inside the boundaries of their cuboid more than $20\%$ of the time in the second half of the episode without visiting the red ball -- since the robotic arm can quickly reach positions, $20\%$ is reasonable; however, the results are not sensitive to this number. 
Similarly, for the second task, we estimated the satisfaction probability using the ratio of the evaluation episodes that visited the red ball either with the green ball or the blue ball, or outside the cuboid of these balls. We did not consider the case where the green and blue balls are jointly visited, as it is not possible in this environment.

Fig. \ref{fig:roboticarm_task0} and \ref{fig:roboticarm_task1} show the results averaged over $4$ simulations for each task. MPG quickly learned a near-optimal controller strategy and successfully converged to an optimal one for the first task. TMPG converged to an optimal strategy slower than MPG. APG learned a near-optimal strategy only after $10^6$ steps. PG could not learn any reasonable strategy. For the second task, MPG immediately learned a near-optimal strategy and then converged to an optimal one. PG converged slower than MPG but faster than TMPG, and TMPG converged faster than APG. Overall, MPG significantly outperformed the other~methods.

}

\section{Conclusions} \label{sec:conc}

In this work, we presented a model-free RL approach to synthesize optimal controller strategies for \textit{any} LTL task in SGs. We provided an approach to craft rewards and discount factors from the parity condition of the DPAs translated from the given LTL tasks. We showed that any controller strategy maximizing the sum of discounted rewards in the worst case also maximizes the minimum probability of satisfying the LTL specification for some sufficiently small parameter. We then introduced our scalable lazy color generation method providing distinct rewards and discount factors only when necessary and thereby improving the learning scalability. In addition, we provided an approximate method that is highly efficient when the controller can perform the task by focusing on a single color.
Finally, we demonstrated the applicability of our methods in several case studies and showed that our methods outperform the existing methods for learning from LTL tasks in SGs.
\update{Generalization of our approach to multi-objectives that include prioritized safety/task constraints (in the form of LTL specifications) and secondary (control-related) cost minimization objectives, as done in \cite{bozkurt2021c} for MDPs, is left as a future work.}

\ifx\arxiv\undefined

\begin{figure*}
    \captionsetup[subfigure]{aboveskip=0pt,belowskip=15pt}
    \centering
    \includegraphics[width=0.7\textwidth]{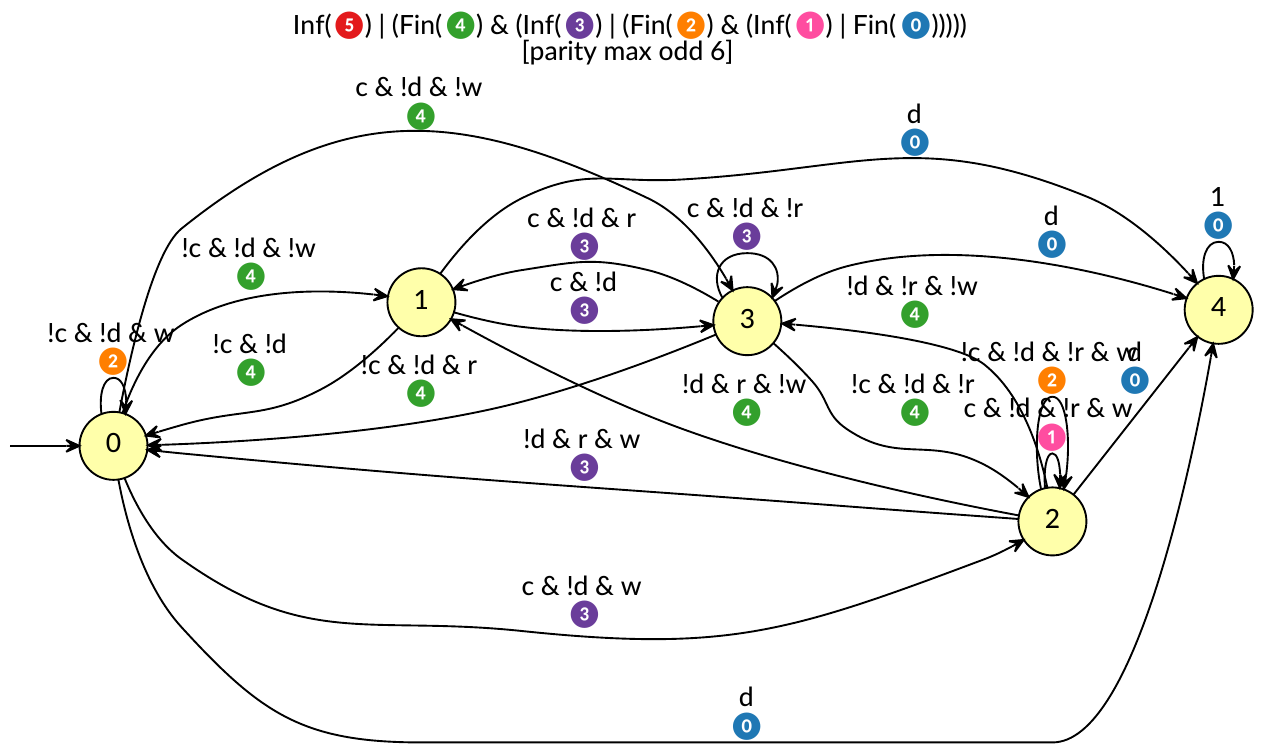}
    \caption{The DPA constructed from the LTL task specification $\varphi_1$ from \eqref{eq:varphi_1} using \emph{Spot} \cite{spot}.}
    \label{fig:dpas}
\end{figure*}
\refstepcounter{subfigure}\label{fig:varphi_1}
\refstepcounter{subfigure}\label{fig:varphi_2}
\refstepcounter{subfigure}\label{fig:varphi_3}
\refstepcounter{subfigure}\label{fig:varphi_4}
\else
\begin{figure*}
    \captionsetup[subfigure]{aboveskip=0pt,belowskip=15pt}
    \centering
    \begin{subfigure}[b]{0.7\textwidth}
       \centering
       \includegraphics[width=\textwidth]{varphi_1.pdf}
       \caption{$\varphi_1$ from \eqref{eq:varphi_1}}
       \label{fig:varphi_1}
    \end{subfigure}
    \hfill
    \begin{subfigure}[b]{\textwidth}
       \centering
       \includegraphics[width=\textwidth]{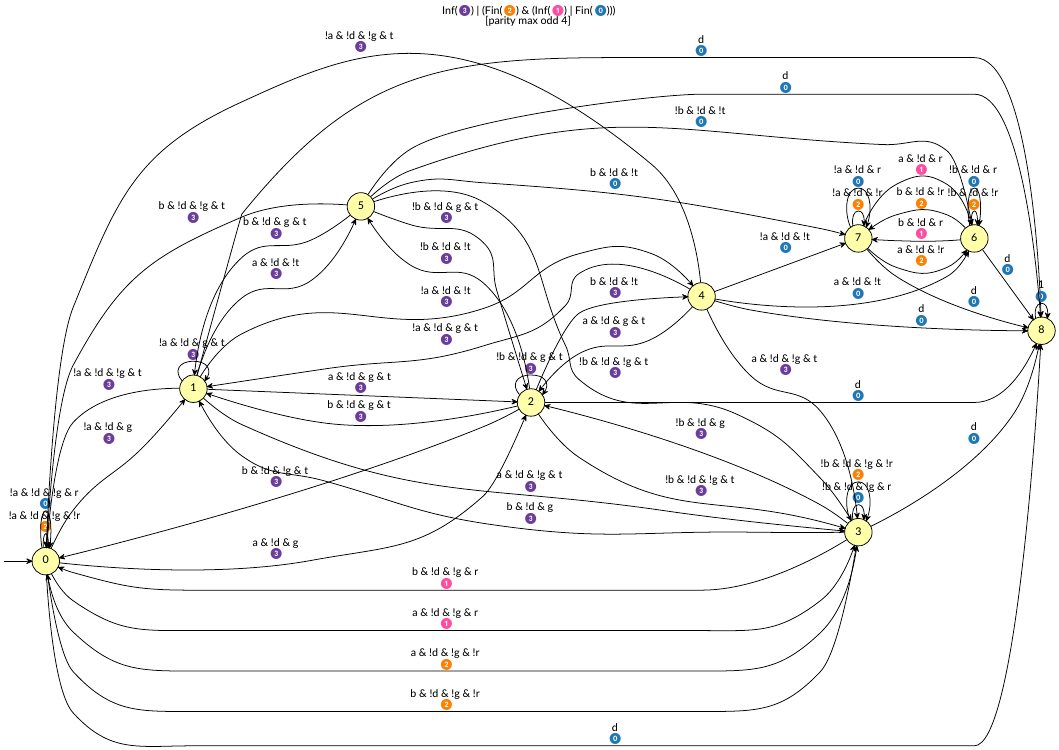}
       \caption{$\varphi_2$ from \eqref{eq:varphi_2}}
       \label{fig:varphi_2}
    \end{subfigure}
\end{figure*}
\begin{figure*}\ContinuedFloat
    \captionsetup[subfigure]{aboveskip=0pt,belowskip=5pt}
    \centering
    \begin{subfigure}[b]{0.35\textwidth}
       \centering
       \includegraphics[width=\textwidth]{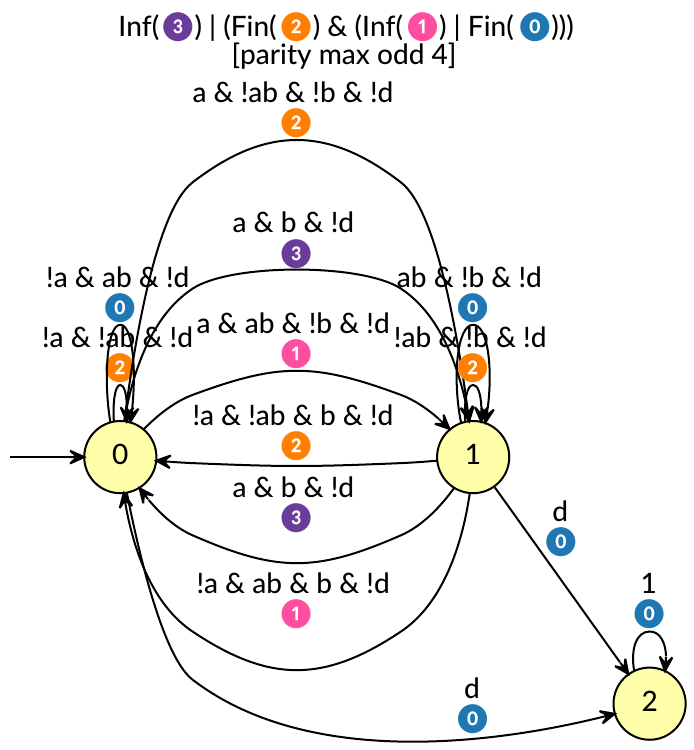}
       \caption{$\varphi_3$ from \eqref{eq:varphi_3}}
       \label{fig:varphi_3}
    \end{subfigure}
    \hspace{50pt}
    \begin{subfigure}[b]{0.3\textwidth}
       \centering
       \includegraphics[width=\textwidth]{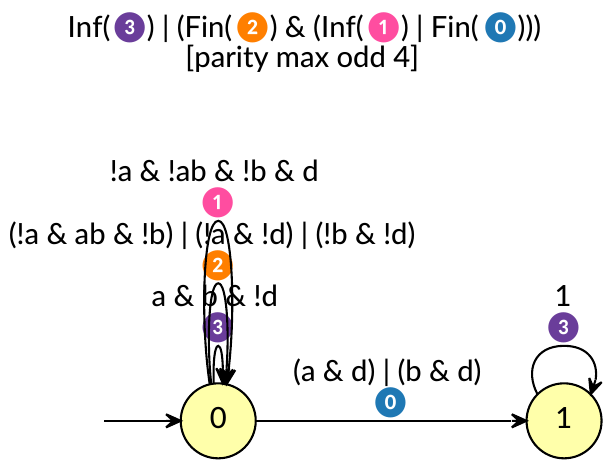}
       \caption{$\varphi_4$ from \eqref{eq:varphi_4}}
       \label{fig:varphi_4}
    \end{subfigure}
    \caption{The DPAs constructed from the LTL task specifications considered in the case studies in Section \ref{sec:case}. The figures are generated using \emph{Spot} \cite{spot}. The circles represent the DPA states; the Boolean formulas and the encircled numbers on the edges represent the labels and the colors of transitions respectively.}
    \label{fig:dpas}
\end{figure*}
\fi

\bibliography{references.bib}
\bibliographystyle{IEEEtran}

\vspace{-20pt}
\begin{IEEEbiography}[{\includegraphics[width=1in,height=1.25in,clip,keepaspectratio]{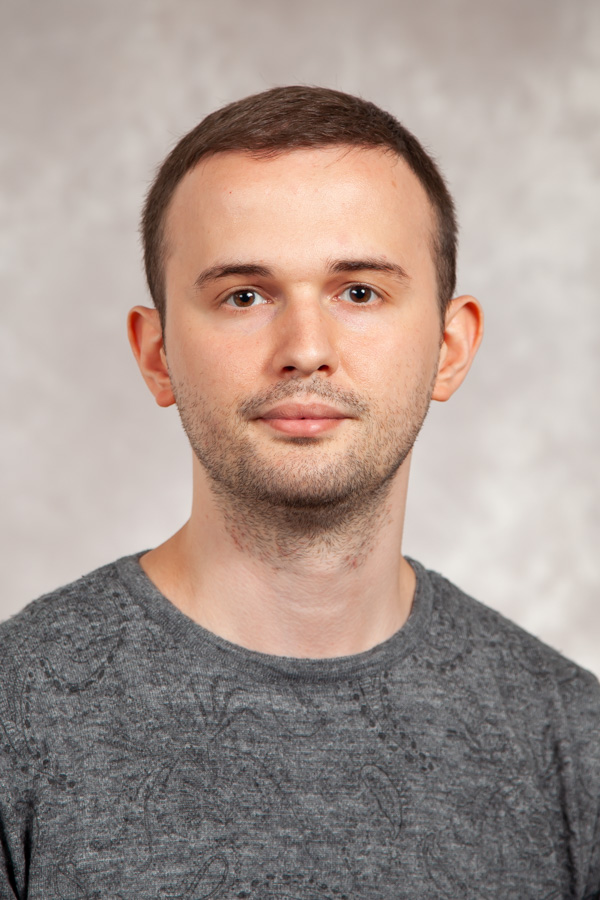}}]{Alper Kamil Bozkurt} received the B.S. and M.S. degrees in computer engineering from Bogazici University, Turkey, in 2015 and 2018, respectively. He is currently a Ph.D. candidate in the Department of Computer Science at Duke University. His research interests lie at the intersection of machine learning, control theory, and formal methods. In particular, he focuses on developing learning-based algorithms that synthesize provably safe and reliable controllers for cyber-physical systems.
\end{IEEEbiography}

\vspace{-20pt}
\begin{IEEEbiography}[{\includegraphics[width=1in,height=1.25in,clip,keepaspectratio]{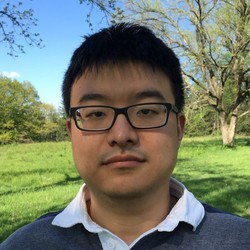}}]{Yu Wang} is an Assistant Professor in the Department of Mechanical and Aerospace Engineering at the University of Florida. He was a postdoctoral associate in the Department of Electrical and Computer Engineering at Duke University. He received his Ph.D. degree in Mechanical Engineering from the University of Illinois at Urbana-Champaign. His research focuses on assured autonomy, cyber-physical systems, machine learning, and formal methods. 
\end{IEEEbiography}

\vspace{-20pt}
\begin{IEEEbiography}[{\includegraphics[width=1in,height=1.25in,clip,keepaspectratio]{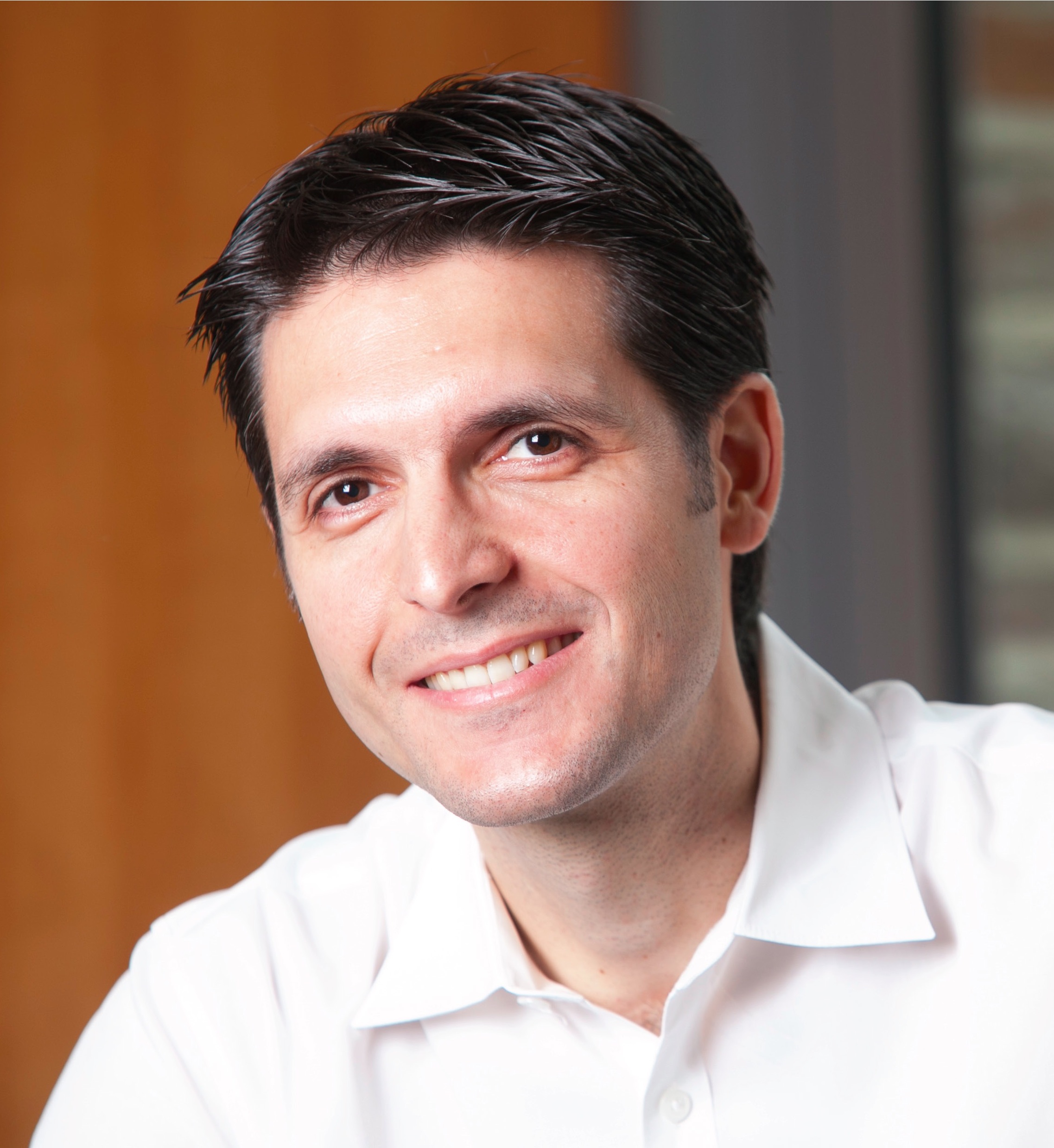}}]{Michael M. Zavlanos} (S'05 M'09 SM'19) received the Diploma in mechanical engineering from the National Technical University of Athens, Greece, in 2002, and the M.S.E. and Ph.D. degrees in electrical and systems engineering from the University of Pennsylvania, Philadelphia, PA, in 2005 and 2008, respectively. He is currently an Associate Professor in the Department of Mechanical Engineering and Materials Science at Duke University, Durham, NC. His research focuses on control theory, optimization, and learning and, in particular, autonomous systems and robotics, networked and distributed control systems, and cyber-physical systems. Dr. Zavlanos is a recipient of various awards including the 2014 ONR YIP Award and the 2011 NSF CAREER Award.
\end{IEEEbiography}

\vspace{-20pt}
\begin{IEEEbiography}[{\includegraphics[trim={ 60pt 230pt 75pt 40pt },,width=1in,height=1.25in,clip,keepaspectratio]{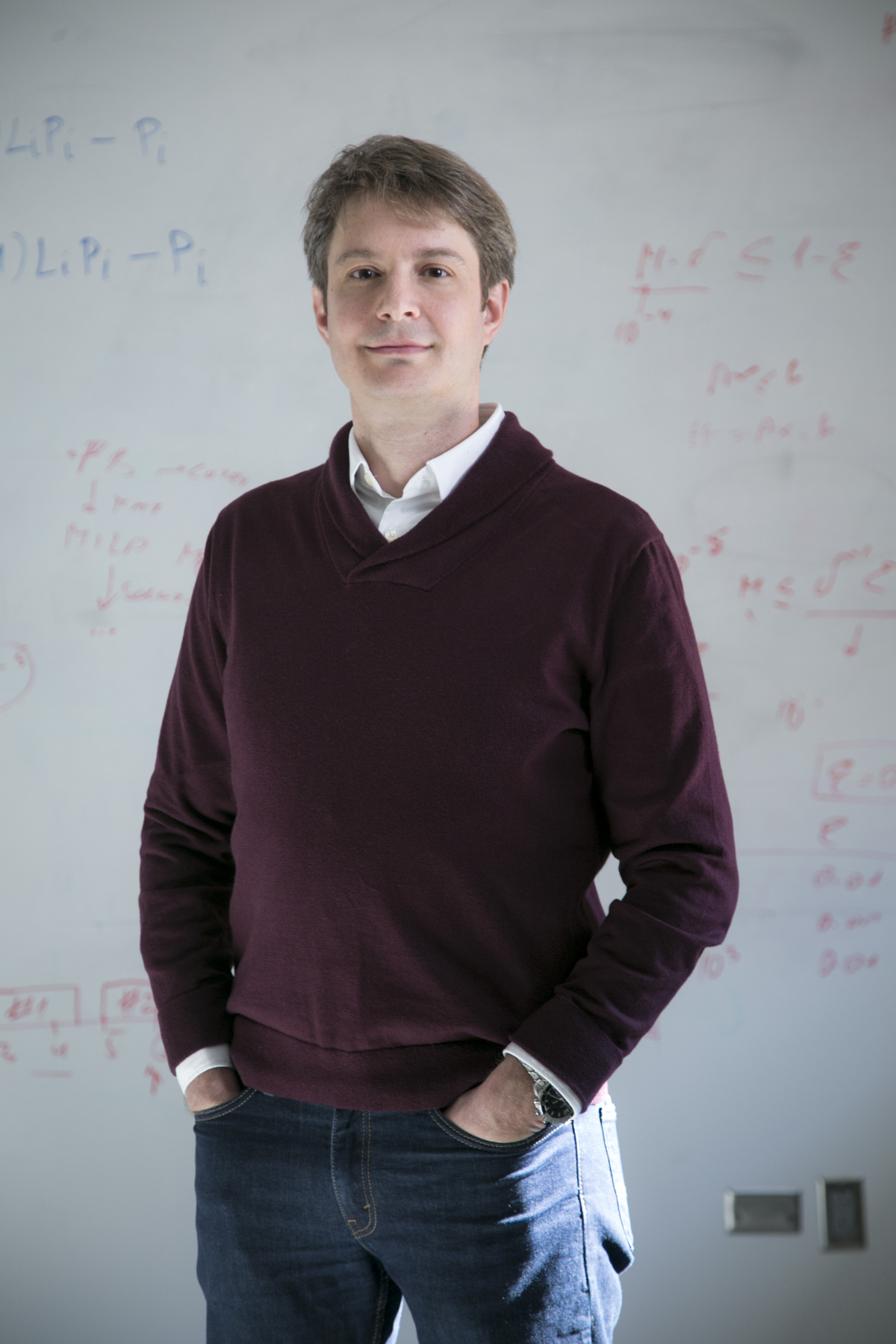}}]{Miroslav Pajic} received the Dipl. Ing. and M.S. degrees in electrical engineering from the University of Belgrade, Serbia, in 2003 and 2007, respectively, and the M.S. and Ph.D. degrees in electrical engineering from the University of Pennsylvania, Philadelphia, in 2010 and 2012, respectively.
He is currently the Dickinson Family Associate Professor in Department of Electrical and Computer Engineering at Duke University. His research interests focus on the design and analysis of high-assurance cyber–physical systems with varying levels of autonomy and human interaction, at the intersection of (more traditional) areas of embedded systems, AI, learning and controls, formal methods, and robotics. 

Dr. Pajic received various awards including the ACM SIGBED Early-Career Award, IEEE TCCPS Early-Career Award, NSF CAREER Award, ONR Young Investigator Award, ACM SIGBED Frank Anger Memorial Award, Joseph and Rosaline Wolf Best Dissertation Award from Penn Engineering, IBM Faculty Award, 
as well as eight Best Paper and Runner-up~Awards. %
\end{IEEEbiography}

\end{document}